\pdfoutput=1
\documentclass[10pt,twocolumn,letterpaper]{article}

\usepackage[pagenumbers]{cvpr} 

\newcommand{\Tab}[1]{\textbf{Tab.~\ref{#1}}} 
\newcommand{\Fig}[1]{\textbf{Fig.~\ref{#1}}}
\newcommand{\Sec}[1]{\textbf{Sec.~\ref{#1}}}
\newcommand{\Eq}[1]{\textbf{Eq.~(\ref{#1})}}

\usepackage{microtype}

\renewcommand{\paragraph}[1]{\vspace{.5em}\noindent\textbf{#1.}}

\DeclareMathOperator*{\argmax}{arg\,max}

\usepackage{amsthm}

\usepackage{soul}

\usepackage{booktabs}
\usepackage{pifont} 
\newcommand{\cmark}{\ding{51}} 
\newcommand{\xmark}{\ding{55}} 
\usepackage{caption}

\usepackage{multirow}

\usepackage[misc]{ifsym}

\usepackage[most]{tcolorbox}

\newtcolorbox[auto counter, number within=section]{takeaway}[1][]{
  enhanced,
  breakable,
  colback=red!4,
  colframe=red!60!black,
  fonttitle=\bfseries,
  title={Takeaway Message \thetcbcounter #1}, 
  #1
}

\definecolor{cvprblue}{rgb}{0.21,0.49,0.74}
\usepackage[pagebackref,breaklinks,colorlinks,allcolors=cvprblue]{hyperref}

\def\paperID{29228} 
\def\confName{CVPR}
\def\confYear{2026}

\title{VL-RouterBench: A Benchmark for Vision–Language Model Routing}

\author{Zhehao Huang$^{1, \dag}$\quad Baijiong Lin$^{2, \dag, \ddag}$\quad Jingyuan Zhang$^{1}$\quad Jingying Wang$^{1}$\\
Yuhang Liu$^{1}$\quad Ning Lu$^{3}$\quad Tao Li$^{1}$ \quad Xiaolin Huang$^{1,\ \textrm{\Letter}}$\\
$^{1}$Institute of Image Processing and Pattern Recognition, Shanghai Jiao Tong University
    \\
$^{2}$The Hong Kong University of Science and Technology (Guangzhou)
\\
$^{3}$The Hong Kong University of Science and Technology
\\
\small{\texttt{\{kinght\_h,tonyzhang666,yuhangliu,highmore,li.tao,xiaolinhuang\}@sjtu.edu.cn}}\\
\small{\texttt{\{blin241\}@connect.hkust-gz.edu.cn}}\quad \small{\texttt{\{ning.lu\}@connect.ust.hk}} \\
}

\begin{document}
\maketitle
\begingroup
\renewcommand{\thefootnote}{}
\footnotetext{$^{\dag}$Equal contribution \ $^{\ddag}$Project lead \ $^{\textrm{\Letter}}$Corresponding author}
\footnotetext{Code \& Dataset: \href{https://github.com/K1nght/VL-RouterBench}{VL-RouterBench}}
\endgroup

\begin{abstract}
Multi-model routing has evolved from an engineering technique into essential infrastructure, yet existing work lacks a systematic, reproducible benchmark for evaluating vision–language models (VLMs). We present \textbf{VL-RouterBench} to assess the overall capability of VLM routing systems systematically. The benchmark is grounded in raw inference and scoring logs from VLMs and constructs quality and cost matrices over sample–model pairs. In scale, VL-RouterBench covers 14 datasets across 3 task groups, totaling 30,540 samples, and includes 15 open-source models and 2 API models, yielding 519,180 sample–model pairs and a total input–output token volume of 34,494,977. The evaluation protocol jointly measures average accuracy, average cost, and throughput, and builds a ranking score from the harmonic mean of normalized cost and accuracy to enable comparison across router configurations and cost budgets. On this benchmark, we evaluate 10 routing methods and baselines and observe a significant routability gain, while the best current routers still show a clear gap to the ideal Oracle, indicating considerable room for improvement in router architecture through finer visual cues and modeling of textual structure. We open-source the complete data construction and evaluation toolchain to promote comparability, reproducibility, and practical deployment in multimodal routing research.

\end{abstract}    
\section{Introduction}
\label{sec:intro}

\begin{figure}[tb]
    \centering
    \vspace*{-12pt}
    \includegraphics[width=\linewidth]{_figs/comprehensive_comparison_best_arxiv.png}
    \caption{\footnotesize{Overall performance comparison on VL-RouterBench. The x-axis and y-axis are Average Cost (Avg. Cost, $\$/10$K samples) and Average Accuracy (Avg. Acc., \%), respectively. Gray dots denote the performance of single models at different costs, and ``Strongest'' and ``Cheapest'' mark the baselines that use only the strongest or the cheapest model. The gray dashed curve depicts the Pareto frontier fitted from these single-model points (only models near the frontier are shown for clarity). ``1st RouterDC'', ``2nd VLC'', and ``3rd MLP'' indicate the top three routers by Rank Score. Points closer to the upper left reflect a better accuracy–cost trade-off. The results show that even advanced routers still have a noticeable performance gap to the Oracle in the VLM routing setting.}}
    \vspace*{-10pt}
    \label{fig:comparison_best}
\end{figure}

\begin{figure*}[tb]
    \centering
    \includegraphics[width=\textwidth]{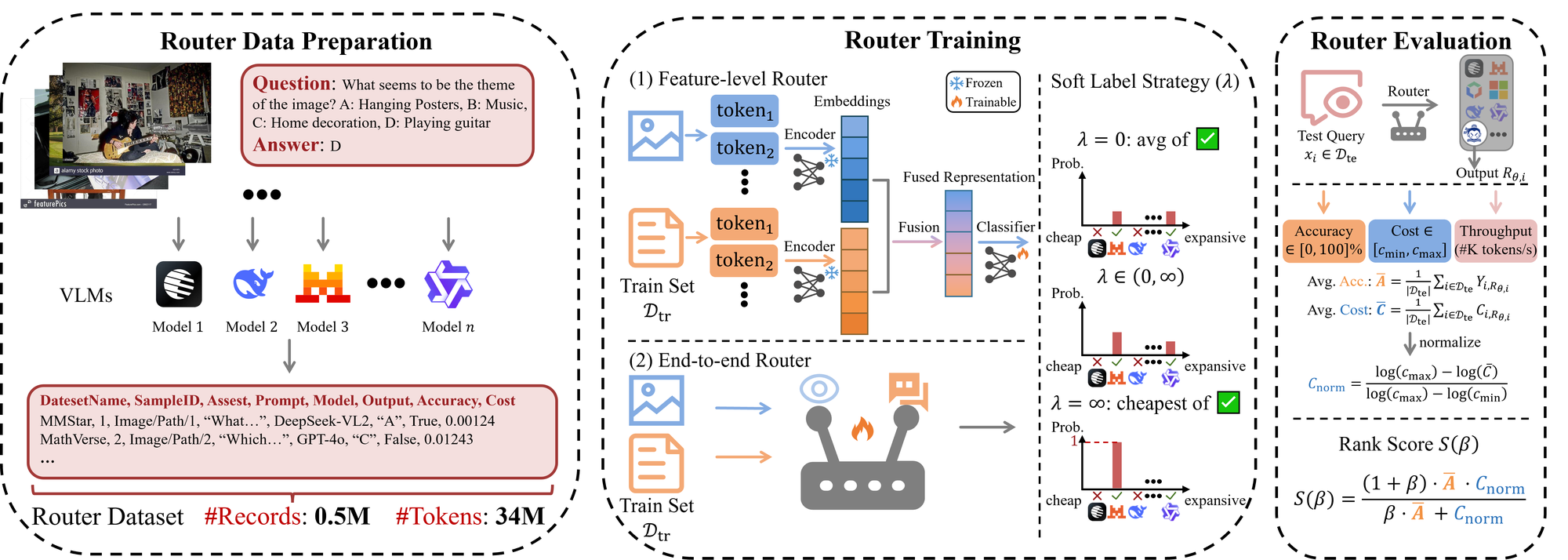}
    \caption{\footnotesize{We propose VL\mbox{-}RouterBench to systematically assess the overall performance of vision-language model routing strategies. The three diagrams on the left, middle, and right represent Router Data Preparation (\Sec{subsec:router_data_preparation}), Router Training (\Sec{subsec:router_training}), and Router Evaluation (\Sec{subsec:router_evaluation_metrics}), respectively.}}
    \vspace*{-6pt}
    \label{fig:vl_routerbench_pipeline}
\end{figure*}
In multi-model systems~\cite{lu2024merge}, the router has gradually evolved from an engineering optimization technique into critical infrastructure. As model families diversify and differ markedly in inference cost and capability, a single model can no longer ensure both performance and efficiency across all request types~\cite{oyelade2025survey}. To address this, a router~\cite{zhang2025beyond} performs dynamic model selection at the query level, enabling flexible trade-offs between quality and cost. The routing ecosystem built around this goal is expanding rapidly. In recent years research~\cite{DBLP:conf/nips/ChenJLK024,DBLP:conf/acl/SongHCGXZ0025,DBLP:conf/acl/ChenWBXW25,DBLP:conf/iclr/FengSY25a} has shown that selecting models at the task or sample level can substantially reduce average inference cost and end-to-end latency while largely preserving output quality, and on this basis more systematic evaluation practices and open benchmarks~\cite{hu2024routerbenchbenchmarkmultillmrouting,huang2025routerevalcomprehensivebenchmarkrouting,feng2025fusionfactoryfusingllmcapabilities,lu2025routerarenaopenplatformcomprehensive} have emerged, advancing the routing paradigm from static rule based configuration to data driven learned router. At the same time, several mainstream commercial model families have embedded routing capabilities within unified product interfaces so that the underlying system can automatically schedule across models, for example the GPT-5~\footnote{\url{https://openai.com/index/introducing-gpt-5}} includes built in routing. Construction and evaluation of effective routers has become a shared need for engineering deployment and academic research.

Although routing research for large language models (LLMs)~\cite{matarazzo2025survey} is maturing, the vision–language model (VLM)~\cite{li2025survey} domain has long lacked a systematic and reproducible routing benchmark. Unlike LLM routing, VLM routing faces more challenges on multiple fronts. First, the task types supported by VLMs are highly diverse, including visual question answering (VQA)~\cite{liu2024mmbench,chen2024we,yue2024mmmu,xai2024grokvision,mathew2022infographicvqa}, visual reasoning~\cite{lu2023mathvista,wang2024measuring,zhang2024mathverse,kembhavi2016diagram}, and chart OCR~\cite{masry2022chartqa,mathew2021docvqa,singh2019towards,liu2024ocrbench}, and different tasks emphasize different aspects of model capability, which makes it more difficult to define what constitutes an ``optimal'' routing decision within a unified framework. Second, the mechanism of multimodal fusion remains an open problem~\cite{bai2025qwen2}. Different VLMs vary substantially in modality interaction and semantic representation, which further increases the difficulty of router design and evaluation. Beyond these challenges, VLM routing also addresses issues introduced by the visual modality, such as the density of visual semantics~\cite{DBLP:conf/mm/RossettoSB24} and cross-modal alignment~\cite{DBLP:conf/icml/RadfordKHRGASAM21}. These factors collectively distinguish VLM routing from LLM routing and constitute its unique difficulties. In view of these differences, we consider it both necessary and urgent to construct a VLM specific routing benchmark that can drive reproducible progress in the research and deployment of VLM routing systems.

We present \textbf{VL-RouterBench} to systematically evaluate the overall capability of VLM routing strategies. To ensure reproducibility and comparability, we design a complete pipeline around router data preparation, router training, and router evaluation, as shown in \Fig{fig:vl_routerbench_pipeline}. First, for data, we use raw inference and scoring logs from VLMs to construct quality and cost matrices. For each sample we retain the image path and textual instruction, the raw answers produced by each candidate model, correctness labels obtained by rule based evaluation as quality, and the counts of input and output tokens, from which we compute the sample level inference cost.  In router training, we employ an adjustable soft label strategy that explicitly controls the trade-off between accuracy and cost, and we consider two routing architectural paradigms, namely feature-level routers and end-to-end routers. In router evaluation, we measure average accuracy, average cost, and throughput for each router on the test set, and we combine normalized cost and average accuracy by a harmonic mean to yield a single rank score, which allows researchers to compare different designs under a unified scale for accuracy and cost. Through large scale data collection, VL-RouterBench covers 14 datasets across three task groups with 30,540 samples, includes 15 open source models and 2 API models, and yields 519,180 sample–model pairs with a total input–output token volume of 34,494,977. 


On VL-RouterBench, we systematically evaluate 10 routing methods and baselines, and further analyze the impact of different text and visual encoders, multimodal feature fusion methods, and multimodal backbones on routing performance, yielding conclusions that are consistent across settings and practically informative. First, we observe a significant routability reward. Under comparable or even lower cost, learned routing systems generally deliver stable gains in accuracy over any single model, which indicates that differences in difficulty and structure among multimodal samples can be effectively captured by lightweight routers. Second, text and visual embeddings followed by simple fusion schemes, such as normalized concatenation, suffices to support highly competitive routers and is consistently superior to counterparts that use only a single modality, which shows that multimodal input provides additional and fine grained discriminative signals for routing decisions. Finally, there remains a noticeable gap between the best current routing methods and the ideal Oracle, which suggests substantial potential for improving router architectures through finer visual cues and through modeling of layout and textual structure, as shown in \Fig{fig:comparison_best}.

We release the full pipeline and the benchmark to ensure that results are reproducible and extensible and that new models and data sources can be plugged in with minimal effort. Our main contributions are summarized below:
\begin{enumerate}
\item We propose VL-RouterBench, a large scale benchmark for VLM routing systems that provides unified data and protocols for comparison of different routing methods.
\item We design a complete pipeline that covers data preparation, router training, and evaluation, and we introduce an adjustable soft label strategy that enables controllable trade-offs between accuracy and cost during training.
\item We conduct comprehensive comparisons and ablations of multiple routing methods on the benchmark, revealing the performance gap between current methods and the ideal Oracle, and offering guidance for future architectural design of VLM routers.
\end{enumerate}

\section{Related Work}
\label{sec:related_work}

\subsection{Routing Systems}

Multi-model routing systems~\cite{lu2024merge} aim to dynamically select the most suitable model for each input to balance quality, cost, and throughput. Early methods~\cite{chen2023frugalgpt}, employed heuristic cascading strategies, invoking models in cost order to significantly reduce inference overhead. Later, systems~\cite{DBLP:conf/emnlp/StripelisX0SJYZ24} modeled routing as the ``quality-cost-throughput" trilemma, validating the comprehensive advantages of learned routing in real-world scenarios. In terms of training methods, existing work can be divided into three categories: supervised learning based on performance logs~\cite{hu2024routerbenchbenchmarkmultillmrouting}, preference-based binary routing~\cite{DBLP:conf/iclr/OngAWC0GKS25}, and soft label training based on reward models~\cite{DBLP:conf/naacl/LuYLLYZZ24,DBLP:conf/nips/ChenJLK024}. Although MoE models~\cite{cai2025survey} also have routing mechanisms, they focus on conditional computations within a single network, which fundamentally differs from multi-model routing at the system level. Current research is mainly focused on LLM routers, and this paper extends the routing paradigm to vision–language scenarios, combining feature-level routing with end-to-end multimodal encoders to explore the impact of cross-modal representations on routing performance.

\subsection{Vision-Language Models}

The rapid development of general VLMs~\cite{li2025survey} provides a rich pool of candidate models for routing. BLIP-2~\cite{DBLP:conf/icml/0008LSH23} provides a solution for cost-effective multimodal modeling by using a lightweight query Transformer, leveraging frozen image encoders and large frozen language models for vision-language pretraining. Building upon this, LLaVA~\cite{li2024llavanext-ablations} introduces visual instruction fine-tuning. InstructBLIP~\cite{DBLP:conf/nips/Dai0LTZW0FH23} systematically investigates cross-task instruction fine-tuning processes and module designs, reflecting the impact of training paradigms on generalization. Recently, the Qwen2.5-VL series~\cite{bai2025qwen2} introduced a dynamic resolution mechanism, assigning adaptive numbers of visual tokens to each image, thereby clearly revealing the trade-off between visual fidelity and computation. In terms of evaluation, \texttt{VLMEvalKit}~\cite{DBLP:conf/mm/DuanYQFCLDZZWL024} integrates hundreds of proprietary and open-source large VLMs along with dozens of benchmark datasets under a unified generative protocol, enabling consistent cross-model comparison and large-scale log collection. Based on this ecosystem, our benchmark collects datasets from three major VLM task groups that are General, STEM, and Charts OCR, and pairs them with VLMs to provide standardized inference pathways for routing data.

\subsection{Router Benchmarks}

Routing evaluation for LLMs has gradually moved towards standardization in recent years. RouterBench~\cite{hu2024routerbenchbenchmarkmultillmrouting} first proposed a systematic routing evaluation framework, providing theoretical analysis and comparisons across various routers based on a unified log format, offering a template and upper-bound estimation protocol for log-driven routing research. Subsequently, RouterEval~\cite{huang2025routerevalcomprehensivebenchmarkrouting} scaled up and systematically reported the model-level scaling-up phenomenon resulting from increased candidate pools and router capabilities meeting the mark, emphasizing the structural benefits of the routing paradigm relative to the single strongest model. To enhance horizontal comparability, RouterArena~\cite{lu2025routerarenaopenplatformcomprehensive} built a routing evaluation dataset that covers a wide range of categories and difficulty levels, accompanied by multi-dimensional metrics including accuracy, cost ratio, optimal selection rate, latency, and robustness, along with an automated leaderboard update process, significantly lowering the barrier for new routing methods to be incorporated and continuously evaluated. All of the above benchmarks focus on text routing, whereas this paper is the first to build a unified evaluation benchmark tailored to the multimodal characteristics of VLMs, providing reproducible training and evaluation protocols.

\section{VL-RouterBench Pipeline}
\label{sec:pipeline}


\subsection{Problem Formulation}
\label{subsec:problem_formulation}


We consider a routing framework for VLMs. Let the sample set be $\mathcal{D}=\{x_i=(I_i,T_i)\}_{i=1}^N$, where $I_i$ denotes an image or a set of images and $T_i$ denotes a textual instruction. The candidate multimodal model set is $\mathcal{M}=\{m_j\}_{j=1}^M$. For any sample $i$, the input is the multimodal pair $x_i=(I_i,T_i)$. A router parameterized by $\theta$ takes $x_i$ as input and produces a decision distribution over models $\pi_\theta(\cdot\mid x_i)$, and during testing the model with the highest probability is selected as the routed model:
\begin{equation}
R_{\theta,i}=R(\theta, x_i)=\argmax\limits_{j\in\{1,\ldots,M\}}\pi_\theta(m_j\mid x_i).
\end{equation}
This formalization enables the router to adaptively assign each input, based on its visual and linguistic content, to the most suitable VLM while balancing performance and efficiency.

To systematically balance routing accuracy and cost, we model router training as a multi objective optimization problem. Let $Y_{i,j}$ and $C_{i,j}$ for each sample–model pair $(i,j)$ denote a performance indicator (such as accuracy) and an inference cost (such as spending dollars). The training objective is
\begin{equation}\label{eq:multi_objective_optimization}
\min_\theta\underbrace{\mathbb{E}\big[-Y_{i,R_{\theta,i}}\big]}_{\text{performance risk}}+\lambda\,\underbrace{\mathbb{E}\big[C_{i,R_{\theta,i}}\big]}_{\text{expected cost}},
\end{equation}
where $\lambda\geq 0$ controls the penalty strength on cost. The objective jointly minimizes performance risk and expected cost to achieve a controllable balance of overall effectiveness, which provides a clear theoretical foundation for subsequent router training methods and supports systematic comparison and analysis of different routing strategies.

\subsection{Router Data Preparation}
\label{subsec:router_data_preparation}

We uniformly extract routing data from the inference and scoring artifacts of \texttt{VLMEvalKit}~\cite{DBLP:conf/mm/DuanYQFCLDZZWL024} to construct a standardized benchmark that ensures fairness and reproducibility of evaluation. To maintain consistency in quality assessment, we adopt a widely used setting in VLM evaluation where each sample prompt contains a single image and correctness can be verified by rule-based criteria, for example multiple choice and answer matching, thereby avoiding biases introduced by multi image inputs or subjective judgment. During data construction we do not consider task group labels of samples. Instead we organize the minimal data unit that records dataset name, sample index, image path, textual prompt, model name, model output, a boolean label for accuracy, and cost of input and output tokens. Using an automated verification procedure that compares model answers with the ground truth, we obtain a correctness label $Y_{i,j} \in \{0,1\}$ for each sample–model pair $(i,j)$ and aggregate all records to form the global quality matrix $Y \in \mathbb{R}^{N \times M}$ for accuracy assessment. Cost is computed from token statistics in actual inference logs together with public model price sheets. Specifically, for sample $i$ and model $j$ the cost is
\begin{equation}\label{eq:cost_calculation}
    C_{i,j}=n^{\text{in}}_{i,j}\cdot c_{j}^{\text{in}}+n^{\text{out}}_{i,j}\cdot c_{j}^{\text{out}},
\end{equation}
where $n^{\text{in}}_{i,j}$ and $n^{\text{out}}_{i,j}$ denote the numbers of input and output tokens respectively, and $c_{j}^{\text{in}}$ and $c_{j}^{\text{out}}$ are the corresponding prices per million tokens for that model. This yields the cost matrix $C \in \mathbb{R}^{N \times M}$ for cost assessment. VL\mbox{-}RouterBench covers 14 datasets with 30,540 samples and involves 15 open source models and 2 API models, resulting in 519,180 complete inference records and a total token count of 34,494,977. The specific datasets and model choices are detailed in \Sec{subsec:datasets} and \Sec{subsec:models}.

\subsection{Router Training}
\label{subsec:router_training}


We divide the entire routing dataset into a training set $\mathcal{D}_{\text{tr}}$, a validation set $\mathcal{D}_{\text{dev}}$, and a test set $\mathcal{D}_{\text{te}}$ with the ratio $|\mathcal{D}_{\text{tr}}| : |\mathcal{D}_{\text{dev}}| : |\mathcal{D}_{\text{te}}| = 7:1:2$. This split is designed to ensure sufficient model training, reliable hyperparameter tuning, and independence of the final evaluation. The validation set is used to optimize the router hyperparameters under preset evaluation metrics so as to strike a reasonable balance between accuracy and cost.

\paragraph{Accuracy–Cost–Aware Soft Label Strategy}
To introduce the trade-off between accuracy and cost into router training, we solve the Lagrangian optimization problem in \Eq{eq:multi_objective_optimization} and obtain an analytic soft label target as
\begin{equation}
t_i^{(\lambda)}(j) = \frac{\mathbf{1}\{Y_{i,j}=1\} \cdot g_\lambda(C_{i,j})}{\sum_{j: Y_{i,j}=1} g_\lambda(C_{i,j})},
\end{equation}
where $\mathbf{1}(\cdot)$ is the indicator function and $g_\lambda(\cdot)$ is a positive function that is monotonically decreasing in cost. Its proof is provided in Appendix~\ref{sec:proof}. Intuitively, the router should place probability mass only on models that answer correctly and should prefer a correct model with lower cost. In practice, we adopt an exponential cost decay $g_\lambda(c)=\exp(-\lambda c)$, which yields a soft label target whose accuracy–cost trade-off is controlled by the hyperparameter $\lambda \ge 0$:
\begin{equation}
t_i^{(\lambda)}(j) = \frac{\mathbf{1}\{Y_{i,j}=1\} \cdot \exp(-\lambda \cdot C_{i,j})}{\sum_{j: Y_{i,j}=1} \exp(-\lambda \cdot C_{i,j})}.
\end{equation}
As shown in \Fig{fig:vl_routerbench_pipeline}, when $\lambda=0$, cost is ignored and probability is distributed uniformly over all correct models, so the router focuses solely on accuracy. As $\lambda \rightarrow +\infty$, the objective strongly favors low cost, and the soft label degenerates to selecting only the cheapest correct model. For $0<\lambda<+\infty$, the objective continuously balances accuracy and cost. We then train the router parameters $\theta$ with the following soft label cross entropy loss:
\begin{align}
\mathcal{L}_\mathrm{soft}(\theta;\lambda)
=\frac{1}{|\mathcal D_{\mathrm{tr}}|}\sum_{i\in\mathcal D_{\mathrm{tr}}}\mathrm{KL}\!\left(t_i^{(\lambda)}\parallel \pi_\theta(\cdot\mid x_i)\right)\\
=\frac{1}{|\mathcal D_{\mathrm{tr}}|}\sum_{i\in\mathcal D_{\mathrm{tr}}}\sum_j t_i^{(\lambda)}(j)\left(-\log \pi_\theta(m_j\mid x_i)\right).
\end{align}

Regarding router architectures, we explore two paradigms. The first uses pretrained encoders to extract text and image features separately and then makes routing decisions with a classifier, as the feature-level router. The second processes multimodal inputs in an end-to-end manner and directly outputs the model selection, as the end-to-end router. Specific implementations are detailed in \Sec{subsec:routing_methods} and Appendix~\ref{sec:detailed_information_about_the_routing_methods}.

\subsection{Router Evaluation \& Metrics}
\label{subsec:router_evaluation_metrics}

To objectively compare different routers, we establish a multi-dimensional evaluation protocol centered on accuracy, cost, and efficiency. Let $R_{\theta,i}$ denote the final model decision made by the router for sample $x_i$. (1) \textbf{Average Accuracy (Avg.\ Acc.)} is
\begin{equation}
\bar A=\frac{1}{|\mathcal D_{\mathrm{te}}|}\sum_{i\in\mathcal D_{\mathrm{te}}}Y_{i,R_{\theta,i}},
\end{equation}
measured in percent and reflecting the overall correctness of routing decisions. (2) \textbf{Average Cost (Avg.\ Cost)} is
\begin{equation}
\bar C=\frac{1}{|\mathcal D_{\mathrm{te}}|}\sum_{i\in\mathcal D_{\mathrm{te}}}C_{i,R_{\theta,i}},
\end{equation}
by default presented as \$/10K samples, which quantifies the economic overhead during inference.
(3) To promote comparability across router configurations, we follow \cite{lu2025routerarenaopenplatformcomprehensive} and introduce a multi-objective \textbf{Rank Score}  that harmonically averages log normalized cost and accuracy. We first apply bounded normalization to cost. Given the full-sample cost range $[c_{\text{min}}, c_{\text{max}}]$, the normalized cost is
\begin{equation}
C_{\mathrm{norm}}=100\times \frac{\log_2(c_{\mathrm{max}})-\log_2\left(\bar C\right)}{\log_2(c_{\mathrm{max}})-\log_2(c_{\mathrm{min}})},
\end{equation}
which is multiplied by $100$ to align with the percentage scale of Avg.\ Acc. We then combine accuracy and normalized cost using the harmonic mean as
\begin{equation}
S(\beta)=\frac{(1+\beta)\cdot{\bar A}\cdot C_{\mathrm{norm}}}{\beta\cdot{\bar A}+C_{\mathrm{norm}}},
\end{equation}
where $\beta>0$ controls the preference between quality and cost. By default we set $\beta=0.1$ to prioritize accuracy.
(4) \textbf{Throughput} measures system efficiency as the number of tokens processed per second in thousands of tokens per second ($\#$K tokens/s), which evaluates the response efficiency of a deployed router system.
(5) In addition to the quantitative metrics mentioned above, we further describe the overall trade-offs at different operating points by plotting accuracy–cost curves. For each given router, all its operating points in the accuracy–cost plane can be fitted into a curve, forming the \textbf{Pareto frontier}. Intuitively, the Pareto frontier closer to the upper left indicates that higher accuracy can be achieved at the same cost, or that lower cost can be achieved at the same accuracy. The detailed process of how to fit the Pareto frontier will be explained in Appendix~\ref{sec:pareto_frontier_fitting}.

\section{Benchmark Curation}
\label{sec:curation}

\subsection{Datasets}
\label{subsec:datasets}

\begin{figure}[tb]
    \centering
    \includegraphics[width=0.8\linewidth]{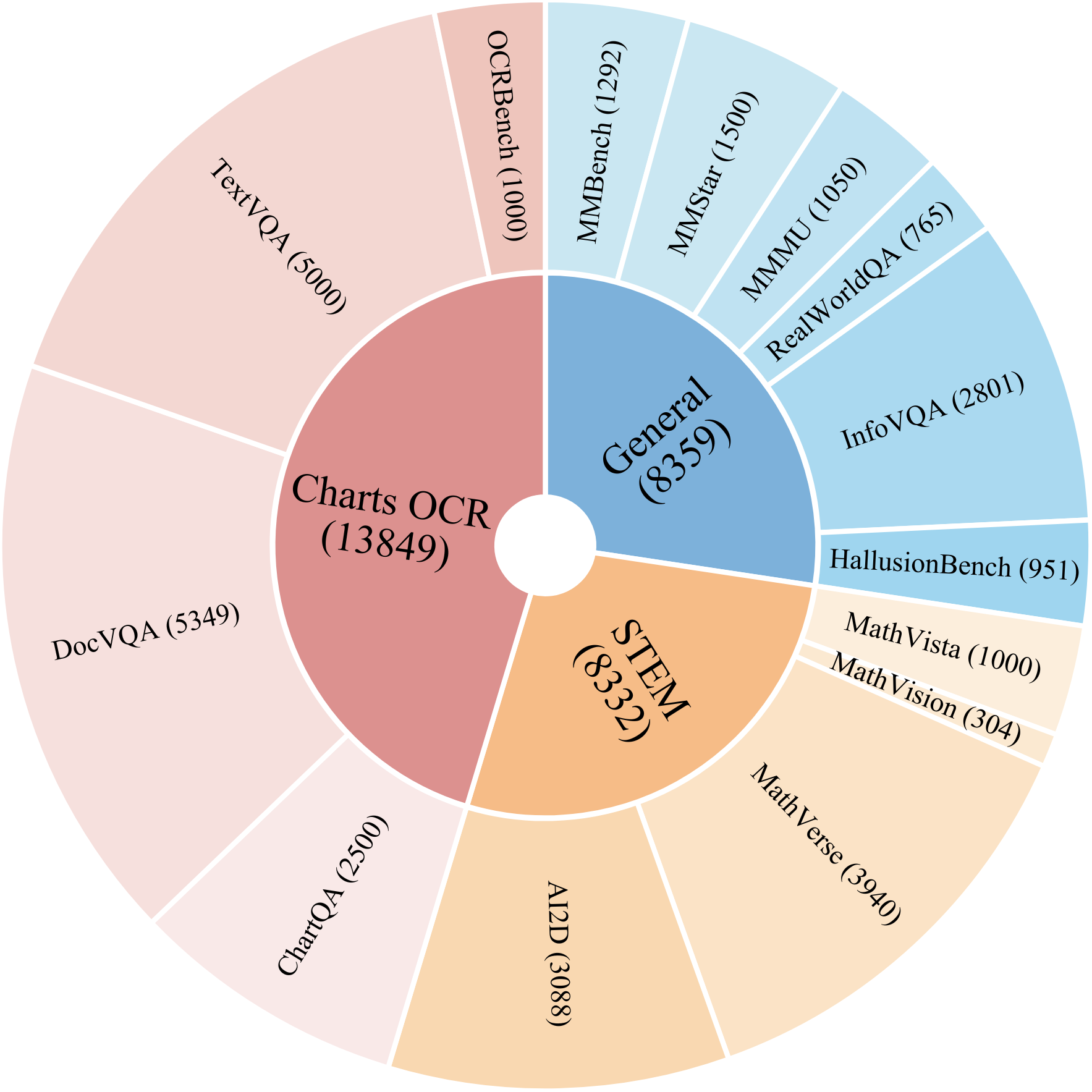}
    \caption{\footnotesize{Dataset distribution in VL\mbox{-}RouterBench. The inner ring shows the three groups, and the outer ring lists the individual datasets. The numbers in parentheses indicate the number of samples contained in each entry.}}
    \vspace{-3pt}
    \label{fig:data_distribution}
\end{figure}

Our benchmark curates 14 datasets across three task groups, namely \textit{General}, \textit{STEM}, and \textit{Charts OCR}, in order to induce sufficient routability differences and to cover diverse real application scenarios. The General group includes MMBench~\cite{liu2024mmbench}, MMStar~\cite{chen2024we}, MMMU~\cite{yue2024mmmu}, RealWorldQA~\cite{xai2024grokvision}, InfoVQA~\cite{mathew2022infographicvqa}, and HallusionBench~\cite{guan2024hallusionbench}, which together balance broad knowledge, evaluate cross domain understanding, assess infographic comprehension, and measure robustness. The STEM group covers MathVista~\cite{lu2023mathvista}, MathVision~\cite{wang2024measuring}, MathVerse~\cite{zhang2024mathverse}, and AI2D~\cite{kembhavi2016diagram}, focusing on structured visual cues and symbolic reasoning in mathematics and scientific diagrams. The Charts and OCR/Document group contains ChartQA~\cite{masry2022chartqa}, DocVQA~\cite{mathew2021docvqa}, TextVQA~\cite{singh2019towards}, and OCRBench~\cite{liu2024ocrbench}, and it systematically examines chart reading and integrated OCR capability. The data distribution is shown in \Fig{fig:data_distribution}, with detailed descriptions of the datasets available in Appendix~\ref{sec:detailed_information_about_the_datasets}. This grouping and unified processing provide a solid foundation for routers to learn generalizable selection boundaries across multiple tasks.

\subsection{Models}
\label{subsec:models}

\begin{table}[t]
    \centering
    \small
    \setlength{\tabcolsep}{5pt}
    \caption{\footnotesize{VLMs used in our benchmark, sorted by parameter count, with per-million-token prices for input and output. For the MoE model, $m$-A-$n$ represents the total number of $m$ B parameters in the model and the number of $n$ B parameters activated during inference. All model prices are estimated based on the model size, compared with the prices on \texttt{Together.ai}.}}
    \vspace{-3pt}
    \label{tab:model_pricing}
    \resizebox{\linewidth}{!}{
    \begin{tabular}{lccc}
    \toprule
    \multirow{2}{*}{\textbf{Model}} & \textbf{Params} & \textbf{Input Price} & \textbf{Output Price} \\
    & \textbf{(B)} & \textbf{(\$/1M tokens)} & \textbf{(\$/1M tokens)} \\
    \midrule
    Janus-Pro-1B~\cite{janus-pro}                  & 1.0   & 0.05 & 0.05 \\
    DeepSeek-VL2-Tiny~\cite{deepseek-vl2}          & 27.0-A-1.0   & 0.05 & 0.05 \\
    SmolVLM2~\cite{marafioti2025smolvlm}           & 2.2   & 0.06 & 0.06 \\
    Kimi-VL-A3B-Thinking-2506~\cite{team2025kimi}  & 16.0-A-2.8   & 0.20 & 0.25 \\
    Phi-3.5-Vision~\cite{abdin2024phi}             & 4.2   & 0.10 & 0.10 \\
    DeepSeek-VL2~\cite{deepseek-vl2}               & 27.0-A-4.5   & 0.35 & 0.50 \\
    Janus-Pro-7B~\cite{janus-pro}                  & 7.0   & 0.18 & 0.25 \\
    MiMo-VL-7B-RL~\cite{Yue2025MiMoVLTR}           & 7.0   & 0.20 & 0.30 \\
    LLaVA-Next-Vicuna-7B~\cite{li2024llavanext-ablations}& 7.0   & 0.20 & 0.20 \\
    Qianfan-VL-8B~\cite{dong2025qianfan}           & 8.0   & 0.18 & 0.25 \\
    Pixtral-12B~\cite{agrawal2024pixtral}          & 12.0  & 0.25 & 0.35 \\
    Gemma3-27B~\cite{team2025gemma}                & 27.0  & 0.35 & 0.50 \\
    Qwen2.5-VL-32B-Instruct~\cite{bai2025qwen2}    & 32.0  & 0.40 & 0.60 \\
    Qwen2.5-VL-72B-Instruct~\cite{bai2025qwen2}    & 72.0  & 0.80 & 1.20 \\
    InternVL2.5-78B~\cite{chen2024expanding}       & 78.0  & 1.00 & 1.50 \\
    Gemini-Flash-2.5~\cite{comanici2025gemini25pushingfrontier} & - & 0.30 & 2.40 \\
    GPT-4o~\cite{hurst2024gpt}                     & -& 2.50 & 10.00 \\
    \bottomrule
    \end{tabular}}
    \end{table}

The core principle in building our model candidate pool is to ensure sufficient heterogeneity across models in terms of scale, architecture, and pricing, in order to realistically simulate the ``quality–cost–latency'' trilemma faced in actual deployment scenarios. To achieve this, we selected 15 open-source models and 2 API models, covering a parameter range from approximately 1B to 78B parameters, and incorporating a variety of mainstream vision encoding and modality alignment paradigms. The selected open-source models include: DeepSeek-VL2~\cite{deepseek-vl2}, DeepSeek-VL2-Tiny~\cite{deepseek-vl2}, Gemma3-27B~\cite{team2025gemma}, InternVL2.5-78B~\cite{chen2024expanding}, Janus-Pro-1B~\cite{janus-pro}, Janus-Pro-7B~\cite{janus-pro}, Kimi-VL-A3B-Thinking-2506~\cite{team2025kimi}, LLaVA-Next-Vicuna-7B~\cite{li2024llavanext-ablations}, MiMo-VL-7B-RL~\cite{Yue2025MiMoVLTR}, Phi-3.5-Vision~\cite{abdin2024phi}, Pixtral-12B~\cite{agrawal2024pixtral}, Qianfan-VL-8B~\cite{dong2025qianfan}, Qwen2.5-VL-32B-Instruct~\cite{bai2025qwen2}, Qwen2.5-VL-72B-Instruct~\cite{bai2025qwen2}, and SmolVLM2~\cite{marafioti2025smolvlm}. The API models include GPT-4o~\cite{hurst2024gpt} and Gemini-Flash-2.5~\cite{comanici2025gemini25pushingfrontier}. 
Model pricing is derived from the publicly available pricing policies of the AI cloud platform \texttt{Together.ai}\footnote{\url{https://www.together.ai/}}, supplemented by estimates based on model size for those not listed. This approach ensures that cost metrics are both comparable and additive across all candidate models. A summary of model parameters and associated pricing is provided in \Tab{tab:model_pricing}, with comprehensive model descriptions available in Appendix~\ref{sec:detailed_information_about_the_models}.

\paragraph{Note} The benchmark pipeline is highly scalable. Its streamlined architecture and modular inference framework enable the easy incorporation of newly released datasets and models. Through a unified feature extraction script, new data and models can be incrementally integrated into the quality–cost matrix, ensuring that this benchmark remains up-to-date and in sync with the rapidly evolving ecosystem of vision-language models and cutting-edge routing research.

\subsection{Routing Methods}
\label{subsec:routing_methods}

\paragraph{Training-Free Baselines} We begin by introducing a set of training-free routing strategies to establish essential performance references for evaluating subsequent data-driven routers. Among them, \textbf{Oracle} establishes the theoretical performance upper bound for any router. It operates with perfect foresight: for each input sample, it selects the lowest-cost model that produces a correct response. If no model can answer the sample correctly, it falls back to the globally cheapest model to minimize cost. Formally, for a given sample $x_i$, we define the set of correct models as $\mathcal{S}_i \triangleq \{j \in \{1, \dots, M\} \mid Y_{i,j} = 1\}$. The Oracle routing decision is given by:
\begin{equation}
R^{\star}(x_i) = 
\begin{cases} 
\arg\min_{j \in \mathcal{S}_{i}} C_{i,j}, & \text{if } \mathcal{S}_{i} \neq \varnothing \\
\arg\min_{j \in \{1, \dots, M\}} C_{i,j}, & \text{if } \mathcal{S}_{i} = \varnothing
\end{cases}.
\end{equation}
\textbf{Strongest} always selects the single model with the highest average accuracy on the entire training set $\mathcal{D}_{\text{tr}}$:
\begin{equation}
R^S(x_i) = \arg\max_{j \in \{1, \ldots, M\}} \frac{1}{|\mathcal{D}_{\mathrm{tr}}|} \sum_{i \in \mathcal{D}_{\mathrm{tr}}} Y_{i,j}.
\end{equation}
Conversely, \textbf{Cheapest} always selects the single model with the lowest average cost over the training set $\mathcal{D}_{\text{tr}}$:
\begin{equation}
R^C(x_i) = \arg\min_{j \in \{1, \ldots, M\}} \frac{1}{|\mathcal{D}_{\mathrm{tr}}|} \sum_{i \in \mathcal{D}_{\mathrm{tr}}} C_{i,j}.
\end{equation}

We now proceed to describe the two learning-based routing paradigms systematically studied in this benchmark.

\paragraph{Feature-level Routers} Feature-level routers extract embeddings $e_{i}^T = E^T(T_i), e_{i}^I = E^I(I_i)$ from the frozen text encoder $E^T$ and visual encoder $E^I$, respectively. The features from both modalities are then aggregated using a fusion function $F(\cdot, \cdot)$ to form a single feature $z_i = F(e_i^T, e_i^I)$. A lightweight classifier is then trained on this feature to predict the model selection distribution $\pi_{\theta}(\cdot \mid x_i) = \text{Cls}(z_i)$. We explore various classifier architectures, including \textbf{KNN}~\cite{shnitzer2023large}, \textbf{PRkNN}~\cite{DBLP:journals/tifs/ZhengZLGZWSL23a}, \textbf{OVR}~\cite{DBLP:journals/jmlr/RifkinK03}, \textbf{K-means}~\cite{shnitzer2023large}, \textbf{Linear}~\cite{shnitzer2023large}, and \textbf{MLP}~\cite{shnitzer2023large}, among others (see Appendix~\ref{sec:detailed_information_about_the_routing_methods} for details). The training objective uses the previously defined accuracy–cost-aware soft label cross-entropy loss. The advantage of this paradigm lies in its excellent portability and training stability, which is well-suited for rapid iteration and feature-level ablation studies on large-scale offline logs, such as replacing visual encoders or adjusting fusion strategies, to better understand the contribution of multimodal representations to routability. Corresponding ablation results are presented and analyzed in \Sec{subsec:ablation_study}.

\begin{figure*}[tbp]
  \centering
  \begin{minipage}[t]{0.59\linewidth}
    \vspace{0pt}
    \setlength{\tabcolsep}{3pt}
    \captionsetup{type=table,aboveskip=0pt,belowskip=3pt}
    \captionof{table}{\footnotesize{Performance of Router Methods on VL-RouterBench. 
    The best and second-best results except for Oracle are highlighted in \textbf{bold} and \underline{underlined}, respectively. 
    \textbf{Avg. Acc.} is average accuracy (\%), \textbf{Avg. Cost} is average cost (\$/10K samples), \textbf{Rank} is sorted by Rank Score, and \textbf{Throughput} is measured in $\#$K tokens/s.}}
    \label{tab:main_results}
    \resizebox{\linewidth}{!}{%
      \begin{tabular}{l|ccccc}
        \toprule
        \textbf{Router} & \textbf{Avg. Acc.} $\uparrow$ & \textbf{Avg. Cost} $\downarrow$ & \textbf{Rank Score} $\uparrow$ & \textbf{Rank} $\downarrow$ & \textbf{Throughput} $\uparrow$ \\
        \midrule
        \multicolumn{6}{c}{\textit{Training-free Baselines}} \\
        Oracle    & 95.60 & 0.37 & 93.68 & 0  & - \\
        Strongest & \underline{78.01} & 2.72 & 68.88 & 8  & - \\
        Cheapest  & 62.43 & \textbf{0.14} & 64.63 & 11 & - \\
        \midrule
        \multicolumn{6}{c}{\textit{Feature-level Routers}} \\
        KNN~\cite{shnitzer2023large}       & 66.26 & \underline{0.38} & 67.13 & 9  & 3.14 \\
        PRkNN~\cite{DBLP:journals/tifs/ZhengZLGZWSL23a}     & 70.68 & 0.41 & 71.09 & 5  & 2.73 \\
        OVR~\cite{DBLP:journals/jmlr/RifkinK03}       & 75.49 & 2.18 & 68.92 & 7  & 6.90 \\
        K-means~\cite{shnitzer2023large}   & 70.01 & 2.21 & 64.62 & 12 & 31.43 \\
        Linear~\cite{shnitzer2023large}    & 68.32\footnotesize{${\pm 7.45}$} & 1.32\footnotesize{${\pm 0.33}$} & 65.68\footnotesize{${\pm 5.69}$} & 10 & \textbf{150.74} \\
        MLP~\cite{shnitzer2023large}       & 77.49\footnotesize{${\pm 0.56}$} & 1.13\footnotesize{${\pm 0.13}$} & 74.23\footnotesize{${\pm 0.22}$} & 3  & \underline{146.71} \\
        \midrule
        \multicolumn{6}{c}{\textit{End-to-End Routers}} \\
        CosineCls~\cite{DBLP:conf/nips/ChenJLK024} & 70.24\footnotesize{${\pm 0.62}$} & 0.61\footnotesize{${\pm 0.16}$} & 69.90\footnotesize{${\pm 0.34}$} & 6 & 7.58 \\
        RouterDC~\cite{DBLP:conf/nips/ChenJLK024}  & 77.52\footnotesize{${\pm 1.04}$} & 1.04\footnotesize{${\pm 0.26}$} & \textbf{74.59\footnotesize{${\pm 1.05}$}} & \textbf{1}  & 6.31 \\
        ZOOTER~\cite{DBLP:conf/naacl/LuYLLYZZ24}   & 74.65\footnotesize{${\pm 1.34}$} & 0.93\footnotesize{${\pm 0.15}$} & 72.58\footnotesize{${\pm 1.03}$} & 4  & 7.25 \\
        VLC~\cite{DBLP:conf/iclr/OngAWC0GKS25}     & \textbf{78.09\footnotesize{${\pm 1.17}$}} & 1.23\footnotesize{${\pm 0.03}$} & \underline{74.33\footnotesize{${\pm 0.51}$}} & \underline{2}  & 6.74 \\
        \bottomrule
      \end{tabular}}
  \end{minipage}\hfill
  \begin{minipage}[t]{0.39\linewidth}
    \vspace{0pt}
    \centering
    \includegraphics[width=\linewidth]{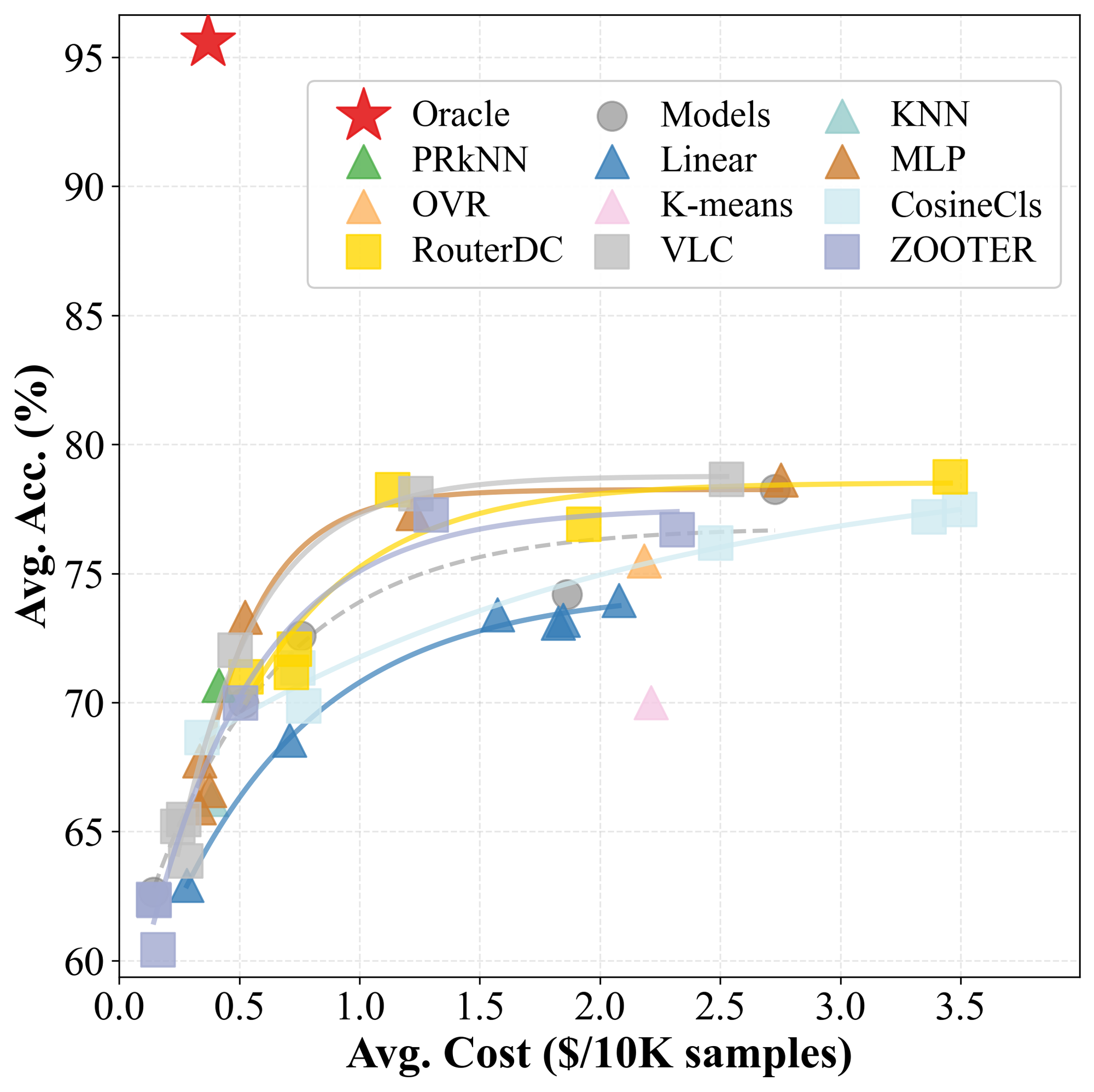}
    \vspace{-14pt}
    \captionsetup{type=figure,aboveskip=0pt,belowskip=0pt}
    \captionof{figure}{\footnotesize{Performance in accuracy–cost plane. Gray dots represent baselines of individual models. Colored markers represent operating points of the same router under different $\lambda$, and curves are Pareto fronts fitted to these points. The closer to the upper left, the better the accuracy-cost trade-off.}}
    \label{fig:comparison_pareto}
  \end{minipage}
\end{figure*}


\paragraph{End-to-End Routers} directly learn the routing decision function from raw images and texts, fine-tuning the entire router network to adapt to specific task combinations. The goal is to capture finer-grained vision–language routing signals on high-difficulty samples. We refer to the design of BERT-Router~\cite{DBLP:conf/iclr/OngAWC0GKS25} in traditional language model routing and explore four vision–language encoders as the backbone for routing: VisualBERT~\cite{li2019visualbert}, UNITER~\cite{DBLP:conf/eccv/ChenLYK0G0020}, LXMERT~\cite{DBLP:conf/emnlp/TanB19}, and ViLBERT~\cite{DBLP:conf/nips/LuBPL19}, constructing a Vision–Language–Classification (\textbf{VLC}) Router and fine-tuning all parameters using the previously mentioned soft label cross-entropy loss. Additionally, we introduce various existing end-to-end routing training strategies, such as \textbf{CosineCls}~\cite{DBLP:conf/nips/ChenJLK024}, \textbf{RouterDC}~\cite{DBLP:conf/nips/ChenJLK024}, \textbf{ZOOTER}~\cite{DBLP:conf/naacl/LuYLLYZZ24}, etc., and evaluate their generalization capabilities in vision–language scenarios. Detailed descriptions of the vision–language encoder structures and routing training strategies can be found in Appendix~\ref{sec:detailed_information_about_the_routing_methods}, with related architecture ablation studies presented in \Sec{subsec:ablation_study}.

\section{Experiments}
\label{sec:experiment}

\subsection{Pilot Study between Oracle and Models}

We first evaluate all candidate models along with the Oracle router on VL-RouterBench, and summarize the outcomes in \Fig{fig:comparison_best} with full results provided in the Appendix~\ref{sec:additional_experiments}. The Oracle, representing the ideal routing strategy, achieves the highest accuracy at a very low cost, highlighting the significant potential of routing systems to enhance the cost-effectiveness of VLMs. Moreover, the substantial accuracy gap between the Oracle and any single model indicates considerable variation in model performance across different samples, underscoring the existence of meaningful routing signals that can be exploited. It is also worth noting that certain open-source models deliver higher accuracy at lower costs compared to other candidates, emphasizing the challenge and opportunity in designing routers that effectively balance accuracy and cost in real-world settings.

\begin{takeaway}
    The large Oracle–single-model performance gap indicates that routing potentially can substantially improve VLM cost-effectiveness compared with any single model.
\end{takeaway}

\subsection{Main Results on Router Methods}

We evaluate all routing methods on VL-RouterBench after training with soft labels that reflect different accuracy-cost trade-offs. Specifically, we consider $\lambda=\{0,10,100,1000,10000,+\infty\}$, and the detailed training settings together with the hyperparameters of each method are provided in Appendix~\ref{sec:additional_experiments}. \Tab{tab:main_results} reports, for each method, the results corresponding to its best Rank Score under the different trade-off settings. Most routing methods surpass the Strongest baseline, which is the single model with the highest accuracy, indicating that routing can indeed achieve high performance at lower cost than any single model. With a multimodal encoder architecture, the improved RouterDC attains the best Rank Score and reaches accuracy close to the strongest single model while using only one fifth of the cost. \Fig{fig:comparison_pareto} presents the accuracy-cost Pareto frontiers of all methods to assess router performance across the global cost span. RouterDC and MLP significantly outperform the other routing methods and achieve similar Pareto optimal fronts. Nevertheless, there remains a substantial gap to the optimal Oracle, which suggests ample room for developing routers with better accuracy-cost trade-offs, while our benchmark provides a reproducible platform for such experimentation.

\begin{figure}[tb]
    \centering
    \includegraphics[width=\linewidth]{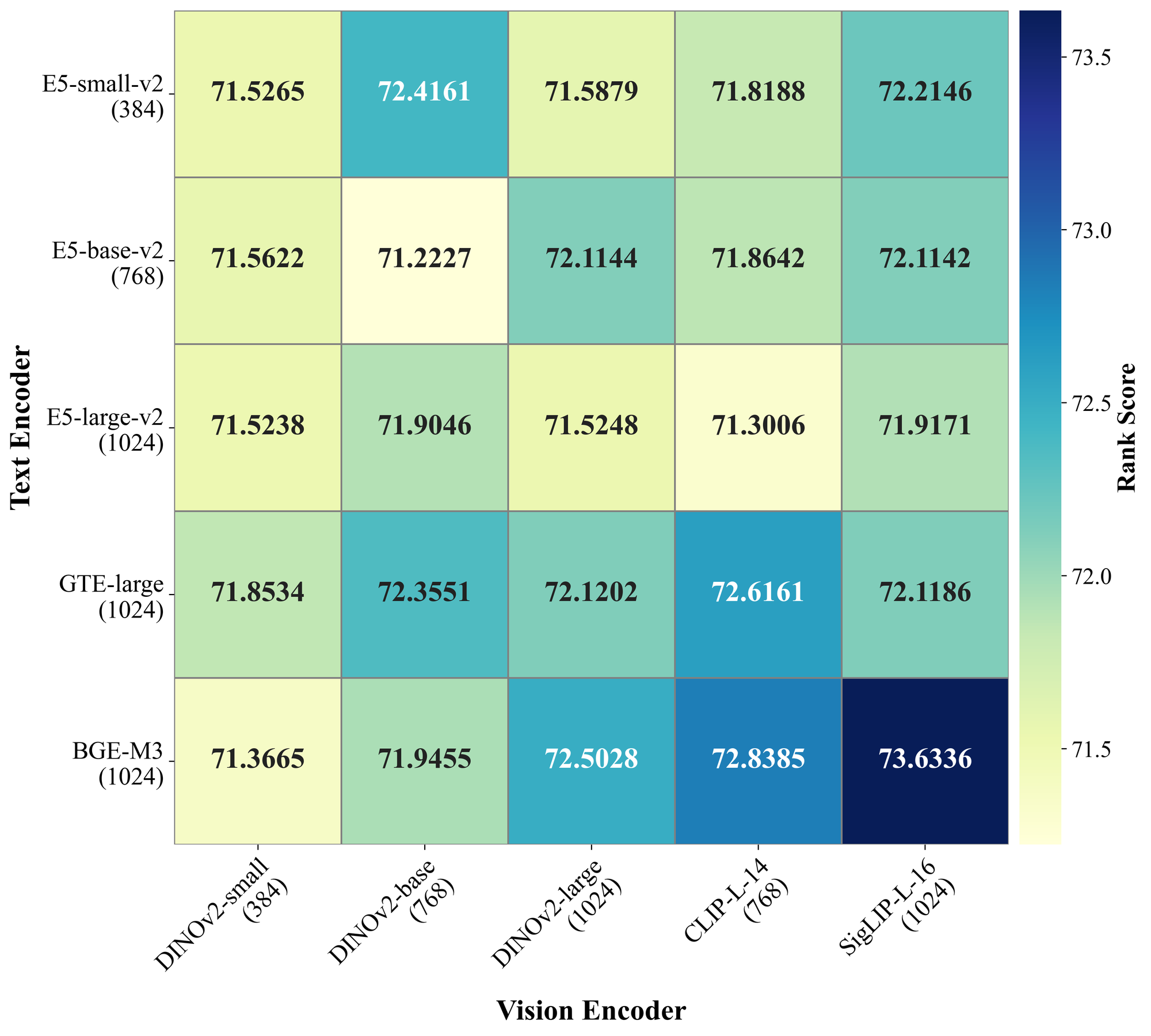}
    \caption{\footnotesize{Performance comparison of different text and visual encoders paired on MLP with $\lambda=100$. The brackets indicate the dimension of the encoder's output embedding. Other settings are the same as in Table~\ref{tab:main_results}.}}
    \vspace*{-3mm}
    \label{fig:encoder_ablation}
\end{figure}

\begin{takeaway}
    Most routers beat the best single-model baseline on the accuracy–cost trade-off, with \textbf{RouterDC} ranking highest. However, all methods still lag well behind the Oracle, showing substantial room to improve.
\end{takeaway}




\subsection{Ablation Study}
\label{subsec:ablation_study}

\paragraph{Text and Visual Encoders}
We conduct ablations on the encoders used to extract text and visual embeddings in the feature level router. Considering different embedding dimensions and architectures, we evaluate 5 text encoders, namely \textbf{E5-small/base/large-v2}~\cite{wang2022text}, \textbf{GTE-large}~\cite{li2023towards}, and \textbf{BGE-M3}~\cite{chen2024bgem3embeddingmultilingualmultifunctionality}, and 5 visual encoders, namely \textbf{DINOv2-small/base/large}~\cite{DBLP:journals/tmlr/OquabDMVSKFHMEA24}, \textbf{CLIP-L-14}~\cite{DBLP:conf/icml/RadfordKHRGASAM21}, and \textbf{SigLIP-L-16}~\cite{DBLP:conf/iccv/ZhaiM0B23}. We pair them in all combinations and train an MLP router. As shown in \Fig{fig:encoder_ablation}, increasing the dimensionality of text and visual embeddings effectively improves the router Rank Score. We adopt the best combination, BGE-M3 and SigLIP-L-16, as the embedding configuration for the feature level router.

\paragraph{Fusion Methods}
We further examine fusion strategies after obtaining text and visual embeddings. We consider using only the text embedding (\textbf{Only-Text}), only the image embedding (\textbf{Only-Image}), simple concatenation (\textbf{Concat}), normalization followed by concatenation (\textbf{Normalize-Concat}), and weighted fusion (\textbf{Weighted-Average}). We also include two methods that introduce additional projection parameters for fine tuning, \textbf{GMU}~\cite{arevalo2017gated} and \textbf{MLB}~\cite{DBLP:journals/tomccap/Roy24}. Formal definitions of these methods are provided in Appendix~\ref{sec:detailed_information_about_the_routing_methods}. Using the fused features, we train an MLP and report results in \Tab{tab:fusion_ablation}. As expected, Only-Text and Only-Image perform weaker because they use information from only one modality. GMU and MLB respectively emphasize accuracy and cost, but they are not competitive in terms of Rank Score. In contrast, Normalize-Concat yields the best accuracy and cost trade-off and is adopted as the default fusion method for all feature level routers. This indicates that there remains ample space to explore multimodal embedding fusion methods tailored for routing.

\begin{table}[tbp]
    \centering
    \small
    \setlength{\tabcolsep}{3pt}
    \caption{\footnotesize{Performance comparison of different fusion methods on MLP with $\lambda=100$. The best and second-best results are highlighted in \textbf{bold} and \underline{underlined}, respectively. Other settings are the same as in Table~\ref{tab:main_results}.} Results are presented as mean and standard deviation across $5$ independent trials.} 
    \label{tab:fusion_ablation}
    \vspace{-3pt}
    \resizebox{\linewidth}{!}{ 
    \begin{tabular}{l|c c c c}
    \toprule
    \textbf{Fusion Method} & \textbf{Avg. Acc.} $\uparrow$ & \textbf{Avg. Cost} $\downarrow$ & \textbf{Rank Score} $\uparrow$ & \textbf{Throughput} $\uparrow$ \\
    \midrule
    Only-Text          & 71.41\footnotesize{${\pm 1.42}$} & 0.60\footnotesize{${\pm 0.11}$} & 70.97\footnotesize{${\pm 1.14}$} & \textbf{350.96} \\
    Only-Image         & 71.89\footnotesize{${\pm 0.61}$} & \underline{0.50\footnotesize{${\pm 0.13}$}} & 71.82\footnotesize{${\pm 0.42}$} & \underline{250.39} \\
    Concat             & 72.35\footnotesize{${\pm 0.93}$} & 0.63\footnotesize{${\pm 0.06}$} & 71.70\footnotesize{${\pm 1.03}$} & 146.71 \\
    Normalize-Concat   & 74.61\footnotesize{${\pm 1.31}$} & 0.55\footnotesize{${\pm 0.11}$} & \textbf{74.05\footnotesize{${\pm 1.06}$}} & 145.12 \\
    Weighted-Average   & \underline{74.78\footnotesize{${\pm 1.02}$}} & 0.64\footnotesize{${\pm 0.10}$} & \underline{73.81\footnotesize{${\pm 1.09}$}} & 146.41 \\
    GMU~\cite{arevalo2017gated}                & \textbf{81.81\footnotesize{${\pm 1.71}$}} & 3.32\footnotesize{${\pm 0.09}$} & 69.12\footnotesize{${\pm 1.44}$} & 141.25 \\
    MLB~\cite{DBLP:journals/tomccap/Roy24}     & 66.43\footnotesize{${\pm 0.14}$} & \textbf{0.30\footnotesize{${\pm 0.04}$}} & 67.66\footnotesize{${\pm 0.57}$} & 141.04 \\
    \bottomrule
    \end{tabular}
    }
\end{table}

\begin{table}[tbp]
    \centering
    \small
    \setlength{\tabcolsep}{3pt}
    \caption{\footnotesize{Performance comparison of different backbones on VLC using $\lambda=100$. 
    The best and second-best results are highlighted in \textbf{bold} and \underline{underlined}, respectively. Other settings are the same as in Table~\ref{tab:main_results}. Results are presented as mean and standard deviation across $5$ independent trials.}}
    \label{tab:backbone_comparison}
    \vspace{-3pt}
    \resizebox{\linewidth}{!}{
    \begin{tabular}{l|cc|cccc}
      \toprule
      \textbf{Backbone} & \textbf{Image} & \textbf{Text} & \textbf{Avg.\ Acc.} $\uparrow$ & \textbf{Avg.\ Cost} $\downarrow$ & \textbf{Rank Score} $\uparrow$ & \textbf{Throughput} $\uparrow$ \\
      \midrule
      BERT~\cite{DBLP:conf/naacl/DevlinCLT19}       & \xmark & \cmark & 63.15\footnotesize{${\pm 0.85}$} & \textbf{0.30\footnotesize{${\pm 0.02}$}} & 64.55\footnotesize{${\pm 0.32}$} & 8.10 \\
      MobileNetV4~\cite{DBLP:conf/eccv/QinLDFLYWBYAAZMH24}  & \cmark & \xmark & \textbf{76.55}\footnotesize{${\pm 1.14}$} & 3.92\footnotesize{${\pm 0.45}$} & 63.33\footnotesize{${\pm 0.72}$} & \textbf{8.93} \\
      VisualBERT~\cite{li2019visualbert} & \cmark & \cmark & 70.54\footnotesize{${\pm 1.12}$} & 0.43\footnotesize{${\pm 0.12}$} & \underline{70.89\footnotesize{${\pm 0.72}$}} & 6.32 \\
      UNITER~\cite{DBLP:conf/eccv/ChenLYK0G0020}     & \cmark & \cmark & 67.69\footnotesize{${\pm 1.10}$} & \underline{0.39\footnotesize{${\pm 0.06}$}} & 68.43\footnotesize{${\pm 0.72}$} & 6.44 \\
      LXMERT~\cite{DBLP:conf/emnlp/TanB19}     & \cmark & \cmark & \underline{71.43\footnotesize{${\pm 1.03}$}} & 0.51\footnotesize{${\pm 0.02}$} & \textbf{71.36\footnotesize{${\pm 0.81}$}} & \underline{6.74} \\
      ViLBERT~\cite{DBLP:conf/nips/LuBPL19}    & \cmark & \cmark & 70.55\footnotesize{${\pm 1.08}$} & 0.45\footnotesize{${\pm 0.10}$} & 70.81\footnotesize{${\pm 0.68}$} & 5.76 \\
      \bottomrule
    \end{tabular}}
  \end{table}

\paragraph{End-to-End Router Backbones}
We next study the impact of the backbone in end-to-end routers. In addition to the 4 multimodal encoders that process both text and images in \Sec{subsec:routing_methods}, we include two single modality baselines: \textbf{BERT}~\cite{DBLP:conf/naacl/DevlinCLT19} from typical LLM router settings using only text, and \textbf{MobileNetV4}~\cite{DBLP:conf/eccv/QinLDFLYWBYAAZMH24} from the vision modality using only images. The performance of different backbones is reported in \Tab{tab:backbone_comparison}. Because they consider only one modality, BERT and MobileNetV4 achieve markedly lower Rank Score than the four multimodal encoders. Among the multimodal encoders the results are very close, and we choose LXMERT, which attains the highest Rank Score, as the backbone for the end to end router. We expect that our benchmark will facilitate the development of multimodal encoder architectures specifically designed for routing to further improve performance.

\begin{takeaway}
    Higher-dimensional text and visual embeddings improve feature-level routing, with the \textbf{BGE-M3 + SigLIP-L-16} pairing performing best, and that simple multimodal fusion via \textbf{Normalize-Concat} delivers the strongest overall Rank Score. For end-to-end routers, multimodal backbones with \textbf{LXMERT} achieving the top performance among closely matched multimodal encoders.
\end{takeaway}

\subsection{Latency Profile}

As the central component for model selection, the router must satisfy low latency and high throughput. The Throughput metric in \Tab{tab:main_results} shows that end to end router methods usually achieve a better accuracy and cost trade-off than feature level routers, yet their throughput is slightly lower due to the complexity of the backbone itself. For feature level routers the primary throughput bottleneck lies in the extraction of feature embeddings. Our benchmark offers a solid platform for studying lighter and higher throughput multimodal architectures specifically designed for routers.

\begin{takeaway}
    End-to-end routers generally achieve better accuracy–cost trade-off than feature-level routers, but their throughput is slightly lower due to the complexity of the backbone itself.
\end{takeaway}
\section{Conclusion}
\label{sec:conclusion}
In this paper, we introduced VL-RouterBench, a systematic and reproducible benchmark for routing in VLM systems. Grounded in raw inference and scoring logs from a pool of 15 open source models and 2 API models over 14 datasets for a total of 519,180 records, the benchmark constructs aligned sample–model quality and cost matrices that capture performance–cost trade-offs at scale and support a unified ranking score. On top of this construction, we defined an evaluation protocol that measures average accuracy, average cost, and throughput, and we evaluated 10 routing methods and baselines, including Oracle upper bounds and non-learned policies. The results reveal a clear routability reward: learned routers consistently outperform any single model at comparable or lower cost, while a sizable gap to the ideal Oracle frontier persists, especially in scenarios that require finer visual cues and more precise modeling of textual structure. We open source the complete pipeline. Looking ahead, VL-RouterBench can be naturally extended to more complex VLM settings, such as multiple image inputs or datasets with subjective or preference-based evaluation, thereby further broadening the application scope of routing and bringing benchmarks closer to real-world multimodal workflows.
\section{Limitations}
\label{sec:limitations}

Here are some potential limitations of VL-RouterBench:

\begin{itemize}
\item \textbf{Single Image Input.} 
The current framework primarily considers single-image input scenarios for VLMs. This design limitation excludes more complex scenarios, such as multiple image inputs, which are relevant for many practical multimodal tasks like image captioning or image-based question answering that involve dynamic and large-scale images. Extending the benchmark to handle multi-image inputs could provide a more comprehensive evaluation of the routing system's capabilities

\item \textbf{Binary Output Decision.} The routing decisions in the VL-RouterBench are based on a binary output model selection (correct or incorrect), while many real-world systems could benefit from more continuous score outputs. A more nuanced approach could involve probabilistic scoring, which would better capture the range of performance variation across models, improving the overall understanding of how well a model performs relative to another in a cost-sensitive manner.

\item \textbf{Limited Evaluation of Cross-Modality.}
Despite the extensive VLM dataset coverage across multiple domains, there is still a significant gap in evaluating cross-modal fusion mechanisms and multi-modal model capabilities. The benchmark primarily evaluates performance based on visual and textual data independently, but real-world applications often require complex fusion between multiple modalities, including audio, video, or even non-textual semantic information. 

\item \textbf{Robustness to Noisy or Imperfect Data.} VL-RouterBench is constructed from standardized logs with clean single-image prompts, and does not explicitly stress-test routers under noisy or imperfect inputs (e.g., typos, paraphrasing, or spurious visual artifacts). As a result, our evaluation primarily reflects routing performance under curated benchmark conditions and may overestimate robustness in real-world deployments where query quality is highly variable. Extending VL-RouterBench with perturbed or user-generated inputs and robustness-oriented metrics is an important direction for future work.

\end{itemize}

\paragraph{Acknowledgment}
The resaerch leading to these results has received funding from the AI for Science Program, Shanghai Municipal Commission of Economy and Informatization (No. 2025-GZL-RGZN-BTBX-02026)  and the National Natural Science Foundation of China (No. 62376155).
{
    \small
    \bibliographystyle{unsrt}
    \bibliography{main}
}

\clearpage
\onecolumn
\setcounter{page}{1}
\setcounter{section}{0}
\setcounter{figure}{0}
\makeatletter 
\renewcommand{\thesection}{\Alph{section}}
\renewcommand{\theHsection}{\Alph{section}}
\renewcommand{\thefigure}{A\arabic{figure}}
\renewcommand{\theHfigure}{A\arabic{figure}}
\renewcommand{\thetable}{A\arabic{table}}
\renewcommand{\theHtable}{A\arabic{table}}

\makeatother
\setcounter{table}{0}
\setcounter{equation}{0}
\renewcommand{\theequation}{A\arabic{equation}}

\maketitlesupplementary


\section{Implications}

VL-RouterBench provides not only a standardized benchmark for VLM routing, but also several broader implications for the design of multimodal systems, routing algorithms, and future evaluation practice.

\begin{itemize}

\item \textbf{Implications for multimodal representation and router architecture.}
Our results show that lightweight multimodal representations (frozen text and image encoders with simple fusion) already support competitive feature-level routers, while end-to-end VLC-style encoders further improve Rank Score at a modest throughput cost. This suggests that routing primarily relies on \emph{coarse but discriminative} multimodal features, motivating router-specific encoders optimized for predicting which downstream model will succeed rather than for full answer generation.

\item \textbf{Cost-aware training as a general recipe.}
By casting routing as a multi-objective optimization over performance and cost and deriving a soft-label target that concentrates probability on correct yet cheaper models, VL-RouterBench provides a generic recipe for cost-aware training. The same Lagrangian-based construction can be reused in other multi-LLM or multi-VLM settings such as tool selection, CoT depth selection, or dynamic resolution control.

\item \textbf{Towards more realistic multimodal routing workloads.}
Built from large-scale VLMEvalKit logs over 14 datasets and 17 models, VL-RouterBench already approximates a realistic mixture of General, STEM, and Charts/OCR workloads, enabling offline prototyping of routing policies and their quality–cost–throughput trade-offs. It also encourages \emph{workload-aware routing}, where router designs and model pools are tailored to domain-specific mixtures of difficulty and modality structure.

\item \textbf{Bridging LLM routing benchmarks and multimodal routing.}
VL-RouterBench extends existing LLM routing benchmarks into the multimodal regime, showing that ideas such as log-based evaluation, deferral curves, optimality analysis, and model-pool scaling transfer naturally but face new challenges from visual tokens and cross-modal alignment. This offers a shared conceptual and tooling foundation for unified routing frameworks across text-only LLMs, VLMs, and future audio or video models.

\end{itemize}

\section{Ethics Statement}
\label{sec:ethics}

\begin{itemize}
\item \textbf{Data Sources and Privacy.}
All data in VL-RouterBench originates from established vision-language benchmarks and the VLMEvalKit framework. We use only publicly available images, texts, and annotations under their respective licenses. The benchmark involves no human data collection, no user interaction, and no attempt to identify individuals. To our knowledge, the underlying datasets contain no personally identifiable information.

\item \textbf{Methodological Scope.}
Our routing approach operates purely at the model selection level and does not modify model internals or training data. While the benchmark itself involves no real-world deployment, we note that selected models may inherit and potentially amplify biases, stereotypes, or harmful behaviors present in their original training corpora.

\item \textbf{Use Restrictions.}
We explicitly prohibit using VL-RouterBench or its components for optimizing routing in harmful applications, including discrimination, misinformation, or surveillance systems. Any practical deployment should incorporate independent safety mechanisms and domain-specific guardrails.

\end{itemize}

\section{Proof of the Soft Label Strategy}
\label{sec:proof}

We consider the routing framework for VLMs described in \Sec{subsec:problem_formulation}.
Let the dataset be $\mathcal{D} = \{x_i = (I_i, T_i)\}_{i=1}^N$ and the candidate model set be $\mathcal{M} = \{ m_j \}_{j=1}^M$.
For each sample $x_i$, a router with parameters $\theta$ produces a distribution $\pi_\theta(\cdot \mid x_i)$ over models and selects
\begin{equation}
R_{\theta,i} = R(\theta, x_i) = \arg\max_{j \in \{1, \dots, M\}} \pi_\theta(m_j \mid x_i).
\end{equation}
For each sample–model pair $(i,j)$, let $Y_{i,j}$ denote a performance indicator (\eg\ correctness) and $C_{i,j}$ denote the inference cost.
The global training objective can be written as the following multi-objective problem:
\begin{equation}
\min_\theta \ \underbrace{\mathbb{E}\left[-Y_{i,R_{\theta,i}}\right]}_{\text{performance risk}}
\ +\ \lambda \, \underbrace{\mathbb{E}\left[C_{i,R_{\theta,i}}\right]}_{\text{expected cost}},
\label{eq:global-multiobjective}
\end{equation}
where $\lambda \ge 0$ controls the strength of the cost penalty.

To derive a cost-aware soft target, we first relax the hard decision $R_{\theta,i}$ to a distribution over models at each sample.
For a fixed sample $x_i$, instead of choosing a single model, we consider a probability vector $q_i = (q_i(1), \dots, q_i(M)) \in \mathbb R^{M}, \sum_{j=1}^M q_i(j) = 1$.
A natural sample-wise relaxation of \Eq{eq:global-multiobjective} is then
\begin{equation}
\min_{q_i \in \mathbb R^{M}}
\sum_{j=1}^M q_i(j)\,\bigl( -Y_{i,j} + \lambda C_{i,j} \bigr),
\label{eq:sample-lagrangian-basic}
\end{equation}
which trades off performance and cost in expectation at $x_i$.

To encourage non-degenerate, smooth distributions, we add an entropy regularizer with temperature $\tau > 0$:
\begin{equation}
\min_{q_i \in \mathbb R^{M}}
\sum_{j=1}^M q_i(j)\,\bigl( -Y_{i,j} + \lambda C_{i,j} \bigr)
\ - \ \tau H(q_i),
\label{eq:sample-lagrangian-entropy}
\end{equation}
where $H(q_i) = - \sum_{j=1}^M q_i(j) \log q_i(j)$ is the Shannon entropy.
We now show that, under the assumption that incorrect models are forbidden, the unique minimizer of \Eq{eq:sample-lagrangian-entropy} coincides with a cost-weighted soft label supported only on correct models.

Fix a sample $x_i$ and let
\begin{equation}
\mathcal{S}_i = \{ j \in \{1,\dots,M\} : Y_{i,j} = 1 \}
\end{equation}
be the set of models that are correct on $x_i$.
Assume $\mathcal{S}_i \neq \varnothing$ and that incorrect models are forbidden, \ie\ $q_i(j) = 0$ whenever $Y_{i,j} = 0$.
Consider the entropy-regularized problem
\begin{equation}
\min_{q_i \in \mathbb R^{M}}
\sum_{j=1}^M q_i(j)\,\bigl( 1 - Y_{i,j} + \lambda C_{i,j} \bigr)
\ - \ \tau H(q_i),
\label{eq:sample-lagrangian-forbidden}
\end{equation}
with $\tau > 0$.
Then the unique minimizer $q_i^\star$ satisfies
\begin{equation}
q_i^\star(j) =
\begin{cases}
\displaystyle
\frac{\exp(-\alpha C_{i,j})}{\sum\limits_{k \in \mathcal{S}_i} \exp(-\alpha C_{i,k})}, & \text{if } j \in \mathcal{S}_i,\\[1.0em]
0, & \text{otherwise},
\end{cases}
\label{eq:soft-target-solution}
\end{equation}
where $\alpha = \lambda / \tau > 0$.
In particular, $q_i^\star$ coincides with the \textbf{cost-aware soft label} $t^{(\lambda)}_{i}(j)$ under $\tau=1$ defined by
\begin{equation}
    t^{(\lambda)}_{i}(j) =
\begin{cases}
\displaystyle
\frac{\exp(-\lambda C_{i,j})}{\sum\limits_{k \in \mathcal{S}_i} \exp(-\lambda C_{i,k})}, & \text{if } Y_{i,j} = 1,\\[1.0em]
0, & \text{otherwise},
\end{cases}
\end{equation}
since only correct models contribute to the denominator.

\begin{proof}
Because incorrect models are useless for training router, we may restrict $q_i$ to be supported on $\mathcal{S}_i$.
Define the restricted simplex
\begin{equation}
\Delta_i^+ = \left\{ q_i:
q_i(j) \ge 0, \ q_i(j) = 0 \ \forall j \notin \mathcal{S}_i \right\}.
\end{equation}
For $j \notin \mathcal{S}_i$ we have $Y_{i,j} = 0$, and under the constraint $q_i(j) = 0$ these terms do not affect the optimization.
On $\mathcal{S}_i$, we have $Y_{i,j} = 1$ and therefore
\begin{equation}
1 - Y_{i,j} = 0, \qquad \forall j \in \mathcal{S}_i.
\end{equation}
Thus, on $\Delta_i^+$ the objective in \Eq{eq:sample-lagrangian-forbidden} simplifies to
\begin{equation}
\tilde{\mathcal L}_i(q_i)
= \sum_{j \in \mathcal{S}_i} q_i(j)\,\lambda C_{i,j}
\ - \ \tau H(q_i),
\label{eq:restricted-objective}
\end{equation}
up to an additive constant that does not depend on $q_i$.
Since $H(q_i)$ is strictly concave and the linear term in $q_i$ is convex, the objective $\tilde{\mathcal L}_i$ is strictly convex on $\Delta_i^+$, and hence has a unique minimizer.

We now compute this minimizer via Lagrange multipliers.
Introduce a scalar multiplier $\mu$ for the normalization constraint $\sum_{j \in \mathcal{S}_i} q_i(j) = 1$.
The Lagrangian is
\begin{equation}
\mathcal{J}(q_i, \mu)
= \sum_{j \in \mathcal{S}_i} q_i(j)\,\lambda C_{i,j}
- \tau H(q_i)
+ \mu ( \sum_{j \in \mathcal{S}_i} q_i(j) - 1 ).
\end{equation}
Taking the derivative with respect to $q_i(j)$ for $j \in \mathcal{S}_i$ and setting it to zero yields
\begin{equation}
\frac{\partial \mathcal{J}}{\partial q_i(j)}
= \lambda C_{i,j}
- \tau (1 + \log q_i(j))
+ \mu
= 0.
\end{equation}
Rearranging gives
\begin{equation}
\log q_i(j)
= \frac{\mu - \lambda C_{i,j}}{\tau} - 1,
\end{equation}
so that
\begin{equation}
q_i(j)
= \exp\!\left(\frac{\mu}{\tau} - 1\right) \exp\!\left(-\frac{\lambda}{\tau} C_{i,j}\right)
= Z_i^{-1} \exp(-\alpha C_{i,j}),
\end{equation}
where we set $\alpha = \lambda / \tau$ and $Z_i = \exp(1 - \mu/\tau)$ is a normalization constant independent of $j$.
Imposing the simplex constraint $\sum_{j \in \mathcal{S}_i} q_i(j) = 1$ determines $Z_i$ as
\begin{equation}
Z_i = \sum_{k \in \mathcal{S}_i} \exp(-\alpha C_{i,k}).
\end{equation}
Therefore the unique minimizer $q_i^\star$ is
\begin{equation}
q_i^\star(j)
=
\begin{cases}
\displaystyle
\frac{\exp(-\alpha C_{i,j})}{\sum\limits_{k \in \mathcal{S}_i} \exp(-\alpha C_{i,k})}, & j \in \mathcal{S}_i, \\[1.0em]
0, & j \notin \mathcal{S}_i,
\end{cases}
\end{equation}
which establishes \Eq{eq:soft-target-solution}.
Finally, note that
\begin{equation}
\sum_{k \in \mathcal{M}} e^{-\alpha C_{i,k}}
= \sum_{k \in \mathcal{S}_i} e^{-\alpha C_{i,k}},
\end{equation}
since all terms with $k \notin \mathcal{S}_i$ vanish after multiplying by the indicator $\mathbf{1}_{Y_{i,k} = 1}$ in the definition of $t^{(\lambda)}_{i}(j)$.
Hence $q_i^\star(j)$ coincides with $t^{(\lambda)}_{i}(j)$ under $\tau=1$, completing the proof.
\end{proof}

\section{Pareto Frontier Fitting}
\label{sec:pareto_frontier_fitting}

To rigorously evaluate the trade-off between inference cost and generation quality, we analyze the Pareto Frontier of the routing system. While RouterBench~\cite{hu2024routerbenchbenchmarkmultillmrouting} proposes a geometric approach based on the Non-Decreasing Convex Hull (NDCH), we implement a parametric exponential saturation model to characterize the diminishing returns of computational investment and estimating the theoretical performance ceiling.

We approximate the Pareto frontier using a smooth exponential saturation function. Let $x$ denote the average cost and $y$ denote the average accuracy. We fit the empirical data to the following non-linear equation:
\begin{equation}
    y(x) = a \cdot (1 - e^{-b \cdot x}) + c
\end{equation}
where $c$ represents the floor performance approximated by the minimum performance in the Pareto set. $a + c$ represents the asymptotic maximum performance that the routing strategy could theoretically achieve given an infinite budget. $b$ is the shape parameter and a larger $b$ indicates high marginal utility in the low-cost regime, characterizing a highly efficient router.

We perform curve fitting using the Trust Region Reflective algorithm to minimize the least squares error. To ensure robust convergence, we initialize the parameters based on the empirical Pareto points, and the goodness of fit is validated using the coefficient of determination.

\section{Detailed Information about the Datasets}
\label{sec:detailed_information_about_the_datasets}

\subsection{General}

\begin{itemize}
\item \textbf{MMBench}~\cite{liu2024mmbench} provides a robust evaluation of 20 fine-grained skills using 3,217 meticulously balanced questions. It introduces a novel CircularEval strategy and uses GPT-4 for answer matching to ensure a comprehensive and robust assessment of multi-modal abilities. 

\item \textbf{MMStar}~\cite{chen2024we} is an elite ``vision-indispensable" multi-modal benchmark comprising 1,500 meticulously human-selected samples, covering 6 core capabilities and 18 detailed axes. The dataset is designed to rigorously evaluate Large VLMs by eliminating samples that can be solved without visual input, thereby addressing data leakage and ensuring true multi-modal competency. 

\item \textbf{MMMU}~\cite{yue2024mmmu} evaluates multimodal models on massive multi-discipline tasks demanding college-level knowledge and deliberate reasoning. It comprises 11.5K questions across six core disciplines and 30 heterogeneous image types, challenging models with expert-level content akin to professional exams.

\item \textbf{RealWorldQA}~\cite{xai2024grokvision} is a benchmark designed for evaluating basic real-world spatial understanding capabilities of multimodal models. The dataset comprises 765 images. 

\item \textbf{InfoVQA}~\cite{mathew2022infographicvqa} benchmarks the understanding of complex infographics, requiring joint reasoning over layout, textual content, and graphical elements. With 30,035 questions across 5,485 images, it specifically challenges models to perform elementary reasoning on diverse data visualizations, bridging the gap between computer vision and document understanding.

\item \textbf{HallusionBench}~\cite{guan2024hallusionbench} stands out as a diagnostic suite tailored to dissect “Language Hallucination $\times$ Visual Illusion" failure modes. Comprising 1,129 handcrafted VQA pairs, it utilizes a unique structure of original versus expert-modified images to test visual dependency rigorously. This approach moves beyond simple accuracy to quantitatively analyze specific failure dimensions.

\end{itemize}

\subsection{STEM}

\begin{itemize}
\item \textbf{MathVista}~\cite{lu2023mathvista} combines diverse mathematical and visual challenges to evaluate fine-grained visual understanding and compositional reasoning. Integrating 28 existing and three new datasets, it covers 6,141 examples across varied visual contexts and seven distinct reasoning types.

\item \textbf{MathVision}~\cite{wang2024measuring} offers 3,040 high-quality visual math problems meticulously selected from 19 math competitions across 12 grades. Strictly cross-validated by experts, it balances open-ended and multiple-choice formats, covering 16 mathematical disciplines to measure reasoning depth.

\item \textbf{MathVerse}~\cite{zhang2024mathverse} assesses genuine visual diagram understanding by transforming problems into six versions with varying multi-modal information. It features 2,612 problems across geometry and functions, employing a novel Chain-of-Thought (CoT) evaluation strategy for fine-grained reasoning assessment.

\item \textbf{AI2D}~\cite{kembhavi2016diagram} addresses diagram interpretation using over 5,000 grade-school science diagrams. It uniquely employs Diagram Parse Graphs (DPG) to model syntactic structure and semantic relationships. With 150,000 rich annotations and 15,000 multiple-choice questions, it challenges models to decode visual symbolism—such as arrows indicating process flow versus consumption—beyond simple object recognition.

\end{itemize}

\subsection{Chart OCR}

\begin{itemize}

\item \textbf{TextVQA}~\cite{singh2019towards} challenges models to read and reason about text embedded within images to answer questions. Comprising 45,336 human-generated questions across 28,408 text-rich images (e.g., billboards, signs) from the Open Images dataset, the benchmark requires integrating optical character recognition (OCR) with semantic reasoning to address complex queries involving scene text.

\item \textbf{ChartQA}~\cite{masry2022chartqa} addresses the limitations of synthetic datasets by evaluating visual and logical reasoning on 20,882 real-world charts collected from diverse online sources. It combines 9,608 human-written questions with 23,111 questions generated from summaries, overcoming issues of fixed vocabularies and template-based queries to support complex reasoning tasks involving aggregation and comparison.

\item \textbf{DocVQA}~\cite{mathew2021docvqa} drives a ``purpose-driven" approach to document analysis, featuring 50,000 questions across 12,767 diverse industry documents (e.g., forms, letters) from the UCSF Library. It challenges models to move beyond simple text extraction, requiring the interpretation of complex layouts, handwriting, and non-textual elements to retrieve information effectively.

\item \textbf{OCRBench}~\cite{liu2024ocrbench} is a comprehensive benchmark tailored to evaluate the OCR capabilities of Large Multimodal Models. It comprises 1,000 manually verified question-answer pairs spanning five diverse components: text recognition, scene-text VQA, document VQA, key information extraction, and handwritten mathematical expressions. This design rigorously tests models on identifying and reasoning with text across varied visual contexts, from artistic fonts to complex documents.

\end{itemize}

\section{Detailed Information about the Models}
\label{sec:detailed_information_about_the_models}

\subsection{Open-source models}

\begin{itemize}

\item \textbf{DeepSeek-VL2}~\cite{deepseek-vl2} is a Mixture-of-Experts vision–language model featuring a total of 27B parameters, with roughly 4.5B parameters activated during inference. It offers a strong and competitive open-source baseline, serving as a higher-capacity counterpart to its Tiny variant. \\
\textit{Official release}: \url{https://github.com/deepseek-ai/DeepSeek-VL2}

\item \textbf{DeepSeek-VL2-Tiny}~\cite{deepseek-vl2} is a cost-efficient MoE variant with 27B total parameters but only 1B activated per token. 
It maintains reasonable multimodal performance while significantly reducing computational cost. \\
\textit{Official release}: \url{https://github.com/deepseek-ai/DeepSeek-VL2}

\item \textbf{Gemma3-27B}~\cite{team2025gemma} is a 27B dense VLM based on the Gemma3 architecture, offering strong general multimodal 
performance and serving as a competitive upper-tier open-source model. \\
\textit{Official release}: \url{https://github.com/google-deepmind/gemma}

\item \textbf{InternVL2.5-78B}~\cite{chen2024expanding} is a 78B high-capacity VLM with advanced vision processing and strong reasoning ability. 
It is the largest model in our open-source pool and enriches the high-end of the Pareto frontier. \\
\textit{Official release}: \url{https://github.com/OpenGVLab/InternVL}

\item \textbf{Janus-Pro-1B}~\cite{janus-pro} is a lightweight 1B VLM designed for cost-sensitive scenarios. Its compact transformer and ViT structure make it suitable as an extremely efficient routing candidate. \\
\textit{Official release}: \url{https://github.com/deepseek-ai/Janus}

\item \textbf{Janus-Pro-7B}~\cite{janus-pro} is a 7B dense VLM equipped with a larger language backbone and a more capable vision encoder than the 1B variant, representing a standard mid-sized open-source model. \\
\textit{Official release}: \url{https://github.com/deepseek-ai/Janus}

\item \textbf{Kimi-VL-A3B-Thinking-2506}~\cite{team2025kimi} is a reasoning-optimized MoE VLM with 2.8B activated parameters. Although efficient, it 
often produces longer output sequences due to its ``thinking-mode'' tuning. \\
\textit{Official release}: \url{https://github.com/MoonshotAI/Kimi-VL}

\item \textbf{LLaVA-Next-Vicuna-7B}~\cite{li2024llavanext-ablations} is a popular LLaVA-style baseline built on Vicuna-7B with a CLIP-like vision encoder. 
It performs well on standard VQA and everyday multimodal tasks. \\
\textit{Official release}: \url{https://github.com/LLaVA-VL/LLaVA-NeXT}

\item \textbf{MiMo-VL-7B-RL}~\cite{Yue2025MiMoVLTR} is a 7B VLM improved with reinforcement learning to encourage more deliberate reasoning. 
It provides an alternative training paradigm compared with purely supervised LLaVA models. \\
\textit{Official release}: \url{https://github.com/InternLM/MiMo}

\item \textbf{Phi-3.5-Vision}~\cite{abdin2024phi} is a vision-augmented version of the Phi-3.5 language model, using a SigLIP-style vision encoder. 
It demonstrates strong efficiency and STEM-oriented reasoning performance. \\
\textit{Official release}: \url{https://github.com/microsoft/Phi-3}

\item \textbf{Pixtral-12B}~\cite{agrawal2024pixtral} is a 12B dense VLM offering competitive performance in document understanding and visual reasoning. 
It provides a middle-high capacity option between 7B and 27B models. \\
\textit{Official release}: \url{https://huggingface.co/mistralai/Pixtral}

\item \textbf{Qianfan-VL-8B}~\cite{dong2025qianfan} is an 8B instruction-tuned VLM from Baidu Qianfan, filling the performance gap between 7B 
and 12B models and enhancing diversity in the mid-range model pool. \\
\textit{Official release}: \url{https://qianfan.cloud.baidu.com}

\item \textbf{Qwen2.5-VL-32B-Instruct}~\cite{bai2025qwen2} is a 32B VLM equipped with resolution-adaptive visual tokenization, making input cost sensitive 
to image size. It provides high accuracy and strong general reasoning. \\
\textit{Official release}: \url{https://github.com/QwenLM/Qwen2.5-VL}

\item \textbf{Qwen2.5-VL-72B-Instruct}~\cite{bai2025qwen2} is a 72B flagship model offering state-of-the-art open-source multimodal reasoning. 
It is one of the most powerful open-source models in our evaluation. \\
\textit{Official release}: \url{https://github.com/QwenLM/Qwen2.5-VL}

\item \textbf{SmolVLM2}~\cite{marafioti2025smolvlm} is a 2.2B compact VLM combining a CLIP-based vision encoder with a lightweight language model. 
It strikes a balance between cost and performance and strengthens the low-to-mid capacity tier. \\
\textit{Official release}: \url{https://huggingface.co/SmolVLM}

\end{itemize}






\subsection{API models}

\begin{itemize}

\item \textbf{Gemini-Flash-2.5}~\cite{comanici2025gemini25pushingfrontier} is a closed-source multimodal model accessed through API, optimized for high throughput and low 
latency. Token usage and cost follow the official API billing. \\
\textit{Documentation}: \url{https://ai.google.dev/gemini-api}

\item \textbf{GPT-4o}~\cite{hurst2024gpt} is a frontier multimodal model from OpenAI providing top-tier performance with relatively low input 
cost but high output-token cost. Token usage is taken directly from API response metadata. \\
\textit{Documentation}: \url{https://platform.openai.com/docs}

\end{itemize}

\section{Detailed Information about the Routing Methods}
\label{sec:detailed_information_about_the_routing_methods}

\subsection{Router Strategy}

\begin{itemize}
\item \textbf{K-Nearest Neighbors (KNN)}~\cite{shnitzer2023large} \\
We implement a similarity-based router using the KNN algorithm, extending the text-based routing approach to the multimodal domain. The core basis is that queries sharing similar visual and textual semantic features should be best served by the same underlying VLM.

\item \textbf{Pairwise Preference Aggregated KNN (PRkNN)}~\cite{DBLP:journals/tifs/ZhengZLGZWSL23a} \\
Unlike the standard KNN router which relies on a single "hard" label per training sample, discarding information about second-best options or close calls, PRkNN utilizes the full correctness matrix and cost matrix of the retrieved neighbors to perform a fine-grained pairwise comparison.

\item \textbf{One-vs-Rest (OVR)}~\cite{DBLP:journals/jmlr/RifkinK03} \\
We implement a decompositional routing strategy based on the One-vs-Rest (OVR) framework. OVR treats the routing problem as a set of $K$ independent binary classification tasks, where $K$ is the number of candidate models in the pool. It assumes that the ``best" model is the one with the highest independent probability of correctness, effectively maximizing the expected accuracy.

\item \textbf{K-means}~\cite{shnitzer2023large} \\
We implement a similarity-based routing approach using $K$ class means, where $K$ is the number of models and a nearest centroid classifier. Instead of storing all training samples, this method aggregates the multimodal features of queries and images best suited for a specific model into a single representative vector. This method significantly reduces inference latency, as it requires only $K$ dot products, rather than searching the entire training database.

\item \textbf{Linear}~\cite{shnitzer2023large} \\
We implement a parametric routing baseline, Linear, which treats the model selection task as a standard multi-class classification problem. Linear learns a global linear decision boundary in the multimodal feature space. Given the fused feature vector $\mathbf{x} \in \mathbb{R}^d$, the router computes a probability distribution $\mathbf{p}$ over the $K$ candidate models via a linear transformation followed by the Softmax function.

\item \textbf{Multi-Layer Perceptron (MLP)}~\cite{shnitzer2023large} \\
We implement a non-linear parametric router using a Multi-Layer Perceptron (MLP), which serves as a more expressive alternative to the linear baseline~\cite{shnitzer2023large}. This method learns a non-linear mapping from the multimodal feature space to the model selection probability distribution, capturing complex interactions between visual and textual semantics.

\item \textbf{Cosine Classifier (CosineCls)}~\cite{DBLP:conf/nips/ChenJLK024} \\
We implement a metric-based routing approach that learns a shared embedding space for both queries and candidate models, leveraging the architecture and Sample-LLM contrastive learning objective proposed in the Cosine Classifier framework~\cite{DBLP:conf/nips/ChenJLK024} to VLM Router. Cosine method fine-tunes a Transformer encoder, to align user queries directly with learnable model prototypes based on semantic suitability.

\item \textbf{Dual-Contrastive Router(RouterDC)}~\cite{DBLP:conf/nips/ChenJLK024} \\
We implement the RouterDC framework to VLM Router, which extends the previously described Cosine Classifier by incorporating a dual-contrastive learning objective. While the Cosine Classifier focuses solely on aligning queries with model prototypes (Sample-LLM alignment), RouterDC introduces an auxiliary Sample-Sample objective to regularize the embedding space by preserving the semantic manifold structure of the queries.

\item \textbf{Zooter}~\cite{DBLP:conf/naacl/LuYLLYZZ24} \\
We implement Zooter, an end-to-end routing framework that fine-tunes a pre-trained Transformer encoder to predict optimal model selection directly from raw text and vision queries, eliminating the need for fixed, pre-extracted embeddings. The router employs a standard Transformer encoder architecture followed by a classification head. Zooter supports both hard and soft label training regimes to balance performance and cost. 

\end{itemize}

\subsection{Fusion Method}
Let the visual embedding be denoted as $\mathbf{v} \in \mathbb{R}^{d_v}$ and the textual embedding as $\mathbf{q} \in \mathbb{R}^{d_q}$. We implement the following fusion strategies to construct a unified representation $\mathbf{f}$ for the routing policy.

\begin{itemize}

\item \textbf{Concatenation} \\
This method preserves the full information from both modalities by joining the vectors along the channel dimension. The fused representation is defined as:
\begin{equation}
    \mathbf{f}_{\text{concat}} = [\mathbf{v}; \mathbf{q}] \in \mathbb{R}^{d_v + d_q},
\end{equation}
where $[\cdot;\cdot]$ represents the concatenation operation.

\item \textbf{Normalized Concatenation} \\
To mitigate scale discrepancies between the visual and textual encoders, we apply $L_2$ normalization to each embedding prior to concatenation:
\begin{equation}
    \mathbf{f}_{\text{norm-concat}} = \left[ \frac{\mathbf{v}}{\|\mathbf{v}\|_2} \,;\, \frac{\mathbf{q}}{\|\mathbf{q}\|_2} \right].
\end{equation}

\item \textbf{Weighted Average} \\
To allow the model to adaptively prioritize one modality (e.g., text-dominant vs. image-dominant queries), we introduce a learnable scalar coefficient $\alpha \in [0, 1]$:
\begin{equation}
    \mathbf{f}_{\text{w-avg}} = \alpha \mathbf{v} + (1-\alpha)\mathbf{q}.
\end{equation}

\item \textbf{Gated Multimodal Unit (GMU)}~\cite{arevalo2017gated} \\
This approach uses a learnable gating mechanism to control the information flow from each modality per sample. We compute a gate vector $\mathbf{z}$ via a sigmoid activation $\sigma$:
\begin{equation}
    \mathbf{z} = \sigma(W_z^v \mathbf{v} + W_z^q \mathbf{q} + \mathbf{b}_z).
\end{equation}
The final fused representation combines the transformed features using $\mathbf{z}$ as a soft switch:
\begin{equation}
    \mathbf{f}_{\text{GMU}} = \mathbf{z} \odot \tanh(W_v \mathbf{v}) + (1 - \mathbf{z}) \odot \tanh(W_q \mathbf{q}).
\end{equation}
where $\odot$ denotes the element-wise product.

\item \textbf{Multimodal Low-rank Bilinear (MLB) Fusion}~\cite{DBLP:journals/tomccap/Roy24} \\
To capture multiplicative interactions between modalities efficiently, MLB projects the inputs into a common space and fuses them via element-wise multiplication (Hadamard product):
\begin{equation}
    \mathbf{h} = \sigma(W_v^T \mathbf{v}) \odot \sigma(W_q^T \mathbf{q}).
\end{equation}
where $\sigma$ is typically the hyperbolic tangent ($\tanh$) function. The final routing prediction is obtained by passing this fused representation through a linear classifier: 
\begin{equation}
    \hat{y} = \text{softmax}(W_{\text{MLP}}\mathbf{h} + \mathbf{b}_{\text{MLP}}).
\end{equation}

\end{itemize}

\subsection{Router Backbone}

We use pretrained transformer models in multiple methods like Cosine Classifier~\cite{DBLP:conf/nips/ChenJLK024}, RouterDC~\cite{DBLP:conf/nips/ChenJLK024}, Zooter~\cite{DBLP:conf/naacl/LuYLLYZZ24}, etc. We select BERT~\cite{DBLP:conf/naacl/DevlinCLT19}, MobileNetv4~\cite{DBLP:conf/eccv/QinLDFLYWBYAAZMH24}, VisualBERT~\cite{li2019visualbert}, UNITER~\cite{DBLP:conf/eccv/ChenLYK0G0020}, LXMERT~\cite{DBLP:conf/emnlp/TanB19} and VILBERT~\cite{DBLP:conf/nips/LuBPL19} as different backbones for experiment and choose the best one to combine with these methods, which turns out to be LXMERT.

\begin{itemize}

\item \textbf{BERT}~\cite{DBLP:conf/naacl/DevlinCLT19} \\
We use BERT as the text-only backbone for our end-to-end router. BERT’s key innovation is its ability to learn representations by conditioning on both left and right context through two pre-training objectives: Masked Language Modeling (MLM), which randomly masks 15\% of input tokens and predicts the original vocabulary based on context, and Next Sentence Prediction (NSP), a binary classification task to predict whether sentence $B$ follows sentence $A$. The special \texttt{[CLS]} token at the beginning of each sequence provides an aggregate representation, $\mathbf{h}_{\text{[CLS]}}$, which is used for classification to predict the optimal model index.

\item \textbf{MobileNetV4}~\cite{DBLP:conf/eccv/QinLDFLYWBYAAZMH24} \\
We use MobileNetV4 as the image-only backbone for our end-to-end router. MobileNetV4's key innovation is the Universal Inverted Bottleneck (UIB) block, which combines the classical Inverted Bottleneck (IB) and ConvNext-style block into a flexible component. The UIB adapts during architecture search, enabling it to function as an IB, ConvNext block, or Feed-Forward Network (FFN) to optimize performance across diverse hardware. To capture long-range dependencies without the overhead of traditional Self-Attention, MobileNetV4 introduces Mobile MQA (Multi-Query Attention), which shares keys and values across attention heads to reduce memory bandwidth demands while preserving global context modeling. This hybrid architecture strikes an excellent balance between latency and accuracy, making MobileNetV4 ideal for real-time routing with minimal inference overhead.

\item \textbf{VisualBERT}~\cite{li2019visualbert} \\
We use VisualBERT as a text-image backbone, representing ``single-stream" multimodal architectures. VisualBERT's key feature is its unified modeling approach, where visual tokens (derived from images) are directly concatenated with text embeddings and processed through a single stack of Transformer layers. This design enables early fusion, allowing deep, self-attention-based interaction between text tokens and image regions at every layer. Compared to late-fusion models, VisualBERT captures fine-grained grounding more effectively, while leveraging pre-trained BERT weights to accelerate convergence.

\item \textbf{UNITER}~\cite{DBLP:conf/eccv/ChenLYK0G0020} \\
We adopt UNITER, a unified Transformer model for joint image-text representation learning. Through pre-training on tasks like Masked Language Modeling, Image-Text Matching, and Word-Region Alignment (via Optimal Transport), it achieves robust cross-modal alignment. UNITER's universal representations enable strong zero-shot or fine-tuned performance on diverse downstream tasks using a single set of weights.

\item \textbf{LXMERT}~\cite{DBLP:conf/emnlp/TanB19} \\
We employ LXMERT as a representative dual-stream architecture for multimodal fusion. It processes vision and language inputs separately before fusing them through a cross-modality encoder. This design enables strong unimodal representation learning and excels in complex reasoning tasks like VQA.

\item \textbf{VILBERT}~\cite{DBLP:conf/nips/LuBPL19} \\
We utilize ViLBERT, a dual-stream architecture that extends BERT to multimodal data. It processes vision and language in separate Transformer streams, interacting through co-attentional layers. These layers use cross-modal attention (e.g., vision queries attend to language keys/values) to enable context-aware fusion while preserving modality-specific representations.

\end{itemize}

\section{Implementation Details}
\label{sec:implementation_details}

All trainable routers in our benchmark share the same data pipeline and optimization protocol, independent of the specific architecture (feature–level routers or end-to-end routers). We start from the VL-RouterBench logs, which provide for each sample $x_i = (I_i, T_i)$ a binary or real–valued quality vector $Y_i \in \mathbb{R}^M$ over the $M$ candidate models and a corresponding cost vector $C_i \in \mathbb{R}^M$. The full quality and cost matrices are denoted by $Y \in \mathbb{R}^{N \times M}$ and $C \in \mathbb{R}^{N \times M}$, where $N$ is the number of samples. We use the pre-defined \texttt{train}/\texttt{dev}/\texttt{test} splits, with the ratio $|\mathcal{D}_{\text{tr}}| : |\mathcal{D}_{\text{dev}}| : |\mathcal{D}_{\text{te}}| = 7:1:2$. 

To supervise the router, we derive either hard or soft training targets over the $M$ experts from $(Y_i, C_i)$ via the multi–objective formulation in \Sec{subsec:router_training}. Concretely, we compute a target distribution that concentrates probability mass on models with high quality and low cost, controlled by a trade-off parameter $\lambda$. We use a set of $\lambda=[0.0, 10.0, 100.0, 1000.0, 10000.0, +\infty]$ in the experiments to control the trade-off between accuracy and cost. For feature-level routers we first compute a fixed text embedding for $T_i$ and then train a shallow classifier on top. For end-to-end routers we attach a task-specific classification head to the final \texttt{[CLS]} or pooled multimodal representation of the visual–language backbone. Unless otherwise specified, we train all routers with AdamW, a learning rate of $2\times 10^{-5}$, batch size $16$, and weight decay $0.01$, using a cosine decay schedule thereafter. We train for at most $5$ epochs. For Linear, MLP, CosineCls, RouterDC, ZOOTER, and VLC, results in main experiments are presented as mean and standard deviation across $5$ independent trials.


\section{Additional Experiments}
\label{sec:additional_experiments}

\subsection{Benchmark Dataset Analysis}

\begin{figure}[tbp]
    \centering
    \includegraphics[width=\linewidth]{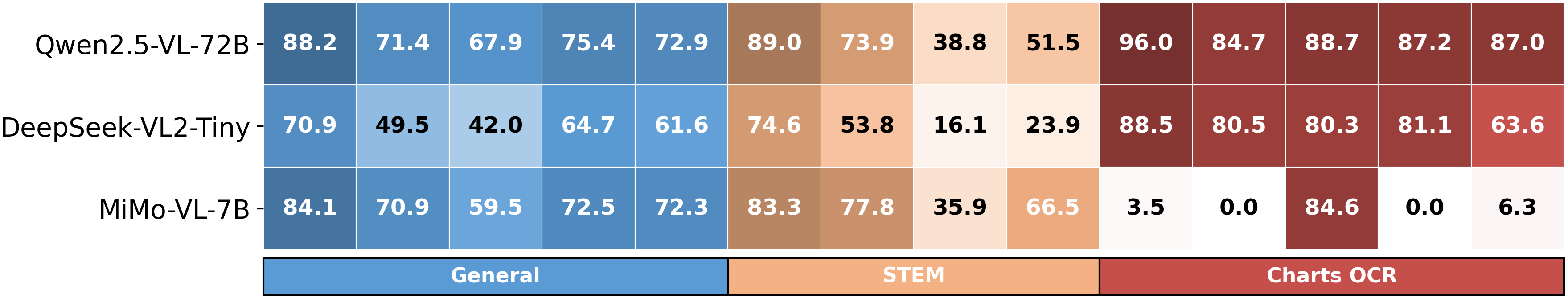}
    \caption{Accuracy heatmap on three representative models, Qwen2.5-VL-72B, DeepSeek-VL2-Tiny, and MiMo-VL-7B. The datasets are grouped into three task groups: General, STEM, and Charts OCR, shown in different colors.}
    \vspace{-3pt}
    \label{fig:benchmark_analysis}
  \end{figure}

We study how accuracy is distributed across task groups/datasets and models. As illustrated by \Fig{fig:benchmark_analysis} on three representative models, VL-RouterBench contains both easy and hard samples rather than being saturated. Importantly, these models show clearly different strength profiles: Qwen2.5-VL-72B is consistently strong but expensive; DeepSeek-VL2-Tiny remains competitive on Charts OCR subsets despite weaker General and STEM; MiMo-VL-7B performs well on General and STEM but collapses on multiple Charts OCR subsets. These results demonstrate meaningful complementarity that routing can exploit. 

\subsection{Ablation Study on $\beta$ in Rank Score}

\begin{figure}[tbp]
    \centering
    \includegraphics[width=\linewidth]{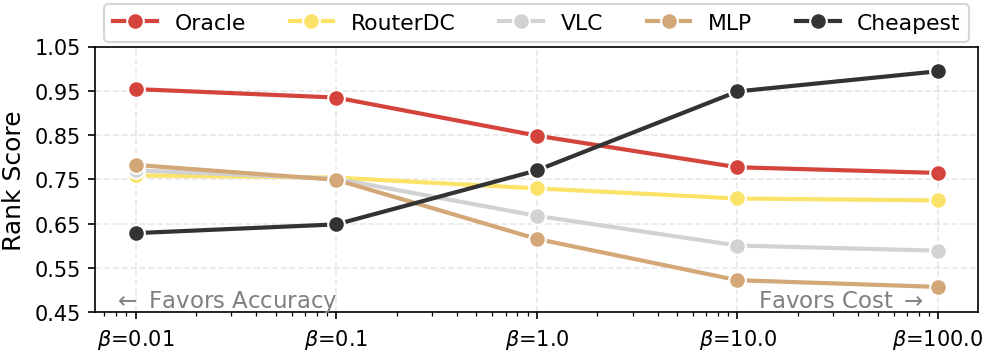}
    \caption{Rank Score comparison of different $\beta$ values on Oracle, RouterDC, VLC, MLP, and Cheapest.}
    \vspace{-3pt}
    \label{fig:ablation_beta}
  \end{figure}

We visualize Rank Score under different $\beta$, including $\beta=0.01, 0.1, 1.0, 10.0, 100.0$, which controls the trade-off between accuracy and cost. \Fig{fig:ablation_beta} shows that Oracle consistently achieves the highest score, while learned routers outperform Cheapest in regimes where accuracy remains important. As $\beta$ becomes large, scores shift predictably toward cost-minimizing routers, causing Cheapest to rise. We note that $\beta$ serves as an application-oriented parameter: users with strict budget constraints may increase $\beta$, whereas those prioritizing quality may decrease it.

\subsection{Extended Multi-cost Framework}

\begin{table}[tbp]
    \centering
    \setlength{\tabcolsep}{3pt}
    \caption{Performance comparison of top routers on VL-RouterBench under multi-cost framework. Other settings are the same as in \Tab{tab:main_results}.}
    \label{tab:multi_cost}
    \vspace{-3pt}
    \resizebox{\linewidth}{!}{
    \begin{tabular}{l|ccccc}
    \hline
    \textbf{Router} & \textbf{Avg. Acc.} $\uparrow$ & \textbf{Avg. Cost} $\downarrow$ & \textbf{Avg. Mem.} $\downarrow$ & \textbf{RS$^*$ $\uparrow$} & \textbf{Rank $\downarrow$} \\
    \hline
    Oracle   & 95.47 & 0.41 & 12.20 & 93.69 & 0 \\
    RouterDC & 70.40 & 0.72 & 16.35 & 70.48 & 1 \\
    VLC      & 65.62 & 0.35 & 11.32 & 68.09 & 2 \\
    MLP      & 65.54 & 0.49 & 12.72 & 67.38 & 3 \\
    \hline
    \end{tabular}
    }
  \end{table}

To better reflect real deployment overhead, we extend VL-RouterBench with an additional cost metric: KV-cache memory (currently available for open-source models). Specifically, we construct a per-sample, per-model KV proxy from (i) logged token counts in inference records and (ii) model architectural KV parameters. Note that this metric is currently computed for open-source models where architectural details are accessible. Building on this, we formalize multi-cost routing as an optimization problem: maximizing expected correctness subject to both financial (\$) and hardware (KV) resource constraints:
\begin{equation*}
\min_{\theta}\; \mathbb{E}\big[-Y_{i,R_{\theta,i}}\big]+ \lambda_C\, \mathbb{E}\big[C_{i,R_{\theta,i}}\big]+ \lambda_{M}\, \mathbb{E}\big[M^{kv}_{i,R_{\theta,i}}\big],
\end{equation*}
which yields a Lagrangian with multiplier $(\lambda_C, \lambda_{M})$. Following the original soft-label derivation, we generalize the target distribution by replacing the single-cost energy with a multi-cost energy: 
\begin{align*}
t^{(\lambda)}_i(j)&=\frac{\mathbf{1}\{Y_{i,j}=1\}\exp(-E_{i,j}(\lambda))}{\sum_{m:Y_{i,m}=1}\exp(-E_{i,m}(\lambda))}, \\
E_{i,j}(\lambda)&=\lambda_C\tilde{C}_{i,j}+\lambda_{M}\tilde{M}^{kv}_{i,j},
\end{align*}
where $\tilde C,\tilde M^{kv}$ are obtained via bounded log scaling for numerical stability and comparability across magnitudes. For evaluation, we use a formal KV-cache memory metric, $M^{kv}$ (MiB), which allows us to report Avg.Mem. alongside the existing Avg.Cost. We also define a Tri-RankScore RS$^*$ by combining $\bar A$, $\tilde C$, and $\tilde M^{kv}$ and using a weighted harmonic mean with ($\beta_C,\beta_M$)=($0.1,0.1$): 
\begin{equation*}
\mathrm{RS}^*(\beta_C,\beta_M)=
\frac{(1+\beta_C+\beta_M)\;\bar{A}\;C_{\text{norm}}\;M_{\text{norm}}}
{
C_{\text{norm}}M_{\text{norm}}
+
\beta_C\,\bar{A}M_{\text{norm}}
+
\beta_M\,\bar{A}C_{\text{norm}}
}.
\end{equation*} 
In experiments under this KV-aware protocol, results in \Tab{tab:multi_cost} show that router comparisons become more deployment-informative and RouterDC attains the best Tri-RankScore among learned routers. Importantly, any further cost metrics can be integrated into our multi-cost optimization framework through the same formalization process.

\subsection{Generalization on Held-out Dataset}

\begin{table}[tbp]
    \centering
    \setlength{\tabcolsep}{3pt}
    \caption{Performance comparison of learned routers on held-out datasets. Routers are trained on 11 seen datasets (\textbf{ID}) and evaluated on 3 held-out datasets (\textbf{OOD}). Other settings are the same as in \Tab{tab:main_results}.}
    \label{tab:generalization}
    \vspace{-3pt}
    \resizebox{\linewidth}{!}{
    \begin{tabular}{l|ccc|ccc}
    \toprule
    \multirow{2}{*}{\textbf{Router}} & \multicolumn{3}{c|}{\textbf{ID}} & \multicolumn{3}{c}{\textbf{OOD}} \\
    \cmidrule(lr){2-4} \cmidrule(lr){5-7}
    & \textbf{Avg. Acc.} $\uparrow$ & \textbf{Avg. Cost} $\downarrow$ & \textbf{Rank Score} $\uparrow$ & \textbf{Avg. Acc.} $\uparrow$ & \textbf{Avg. Cost} $\downarrow$ & \textbf{Rank Score} $\uparrow$ \\
    \midrule
    Oracle   & 94.85 & 0.39 & 92.87 & 96.78 & 0.47 & 93.94 \\
    RouterDC & 73.62 & 0.53 & 73.26 & 68.81 & 0.54 & 68.88 \\
    VLC      & 72.09 & 0.49 & 72.07 & 69.36 & 0.52 & 69.44 \\
    MLP      & 71.61 & 0.73 & 70.66 & 55.53 & 0.86 & 55.79 \\
    \bottomrule
    \end{tabular}
    }
  \end{table}

To investigate generalization of learned routers, we introduce a held-out-dataset protocol: we select one unseen dataset per task group (default: MMMU for General, MathVision for STEM, and ChartQA for Charts/OCR), train routers on the remaining 11 datasets, and evaluate on ID (test splits of the 11 seen datasets) versus OOD (test split of the 3 held-out datasets only). We then report the ID and OOD Avg. Acc., Avg. Cost, and Rank Score. Under this stricter setting, learned routers remain competitive on OOD and preserve meaningful behavior rather than collapsing, e.g., RouterDC achieves 68.88 OOD Rank Score generalized from 73.26 ID Rank Score. Results are shown in \Tab{tab:generalization}.

\subsection{Oracle-Best Gap Study}

\begin{figure}[tbp]
    \centering
    \vspace{-3mm}
    \includegraphics[width=\linewidth]{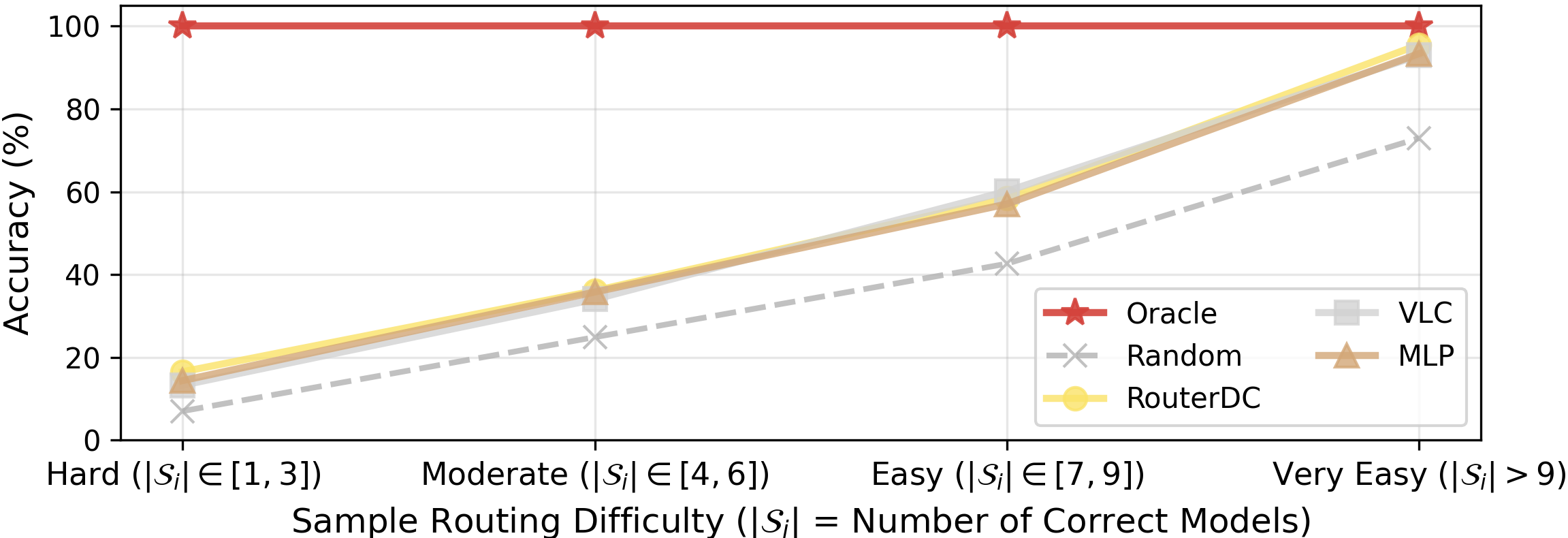}
    \caption{Accuracy distribution on different routing difficulty levels for top routers. $|\mathcal S_i|$ is the number of models correctly answering sample $i$.}
    \vspace{-3pt}
    \label{fig:sample_routing_difficulty}
  \end{figure}

To analyze the source of the Oracle-Best gap, we focus on routing-sensitive samples. We stratify instances by $|\mathcal S_i|$ (the number of models correctly answering sample $i$), which serves as a proxy for routing difficulty (smaller $|\mathcal S_i|$ means harder routing). As shown in \Fig{fig:sample_routing_difficulty}, while top routers consistently outperform Random baseline, their performance drops sharply as difficulty increases, leaving substantial room for improvement relative to Oracle on hard-route samples. This indicates that the Oracle–Best gap is primarily driven by routers’ limited ability to extract sufficiently discriminative multimodal signals when only a few specialized models can solve the instance. These hard-route cases therefore keep our benchmark challenging and motivate future work on stronger multimodal routing.

\subsection{Results on Oracle and Models}

\begin{figure*}[tb]
    \centering
    \includegraphics[width=\linewidth]{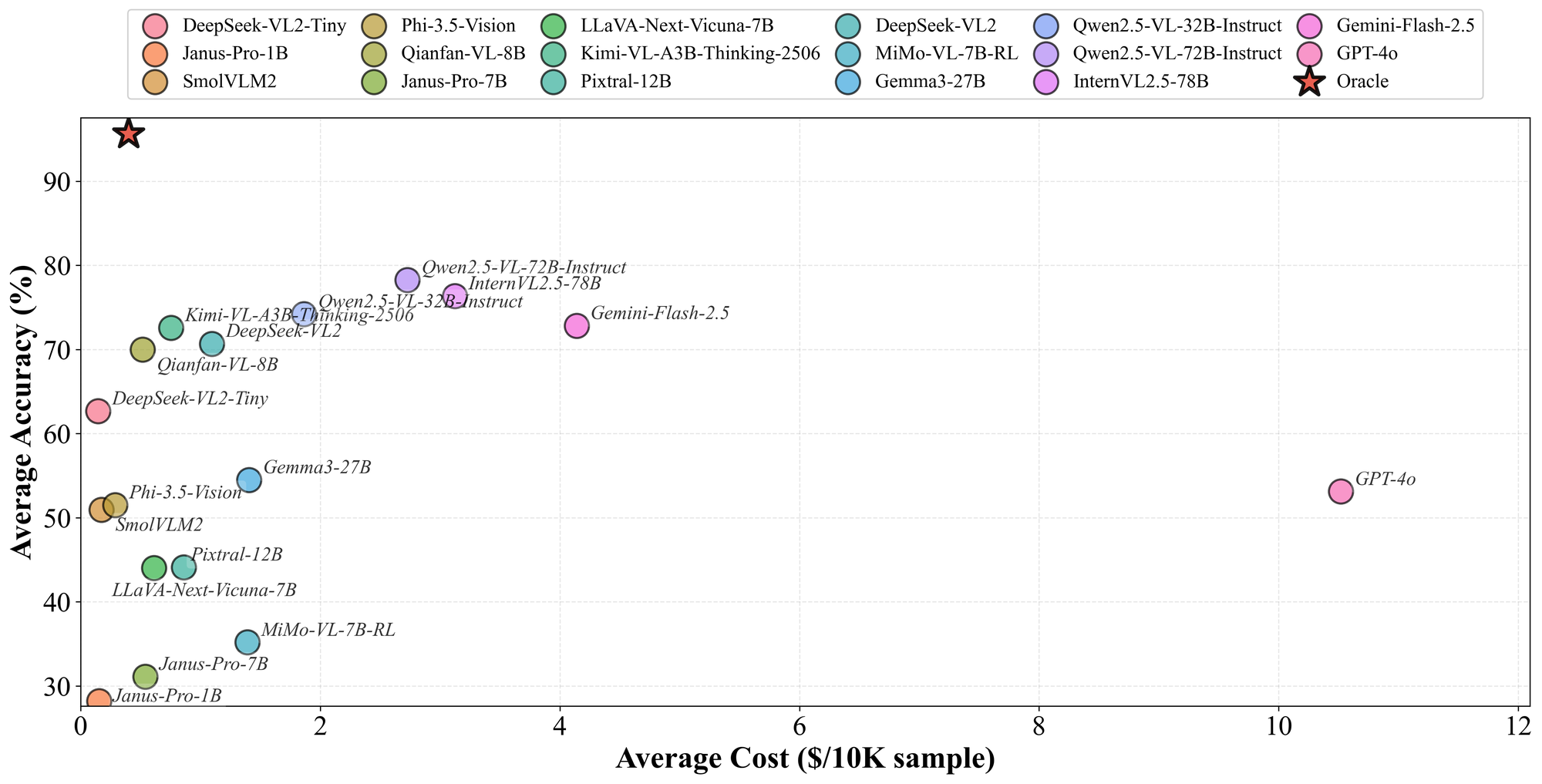}
    \caption{Accuracy-Cost Curve of Oracle and Models. The closer to the upper left corner, the better the accuracy-cost trade-off.}
    \vspace*{-3mm}
    \label{fig:oracle_and_models}
\end{figure*}

We provide the detailed results of all candidate models and the Oracle router on VL-RouterBench in \Fig{fig:oracle_and_models}. The Oracle represents an ideal routing policy and achieves the highest average accuracy at very low average cost, which highlights the substantial potential of routing systems for improving the cost effectiveness of VLM deployments. It is also noteworthy that the two API models, despite their significantly higher prices, do not obtain better average accuracy than several cheaper open source candidates. This pattern poses a challenge for practical router design, since an effective routing strategy must explicitly avoid high cost options whose accuracy does not justify their price and instead maintain a favorable balance between accuracy and cost.

\subsection{Detailed Results on Each Dataset}


We further report per-dataset detailed results under different values of $\lambda$, as shown in \Tab{tab:dataset_results_arxiv_lambda_0.0}-\Tab{tab:dataset_results_arxiv_lambda_inf}. These tables characterize the accuracy-cost trade-offs achieved by different routing methods across a wide range of $\lambda$, providing a fine-grained view of how each method adapts under varying cost preferences.

\newpage
\newpage 
\begin{table*}[tbp]
    \centering
    \small
    \setlength{\tabcolsep}{3pt}
    \caption{Performance comparison of router methods on VL-RouterBench across datasets for $\lambda=0.0$. \textbf{Avg. Acc.} is average accuracy (\%), \textbf{Avg. Cost} is average cost (\$/10K samples). The best and second-best results among learned routers are highlighted in \textbf{bold} and \underline{underlined}, respectively (excluding Oracle and baseline methods).}
    \label{tab:dataset_results_arxiv_lambda_0.0}
    \vspace{-3pt}
    \resizebox{\linewidth}{!}{
    \begin{tabular}{c c c|c c c c c c c c c c c c c}
    \toprule
    \multirow{2}{*}{\textbf{Group}} & \multirow{2}{*}{\textbf{Dataset}} & \multirow{2}{*}{\textbf{Metric}} & \multicolumn{3}{c}{\textbf{Baselines}} & \multicolumn{4}{c}{\textbf{Hard Label}} & \multicolumn{6}{c}{\textbf{Soft Label ($\lambda=0.0$)}} \\
    \cmidrule(lr){4-6} \cmidrule(lr){7-10} \cmidrule(lr){11-16}
     &  &  & \textbf{Oracle} & \textbf{Strongest} & \textbf{Cheapest} & \textbf{KNN} & \textbf{PRkNN} & \textbf{OVR} & \textbf{K-means} & \textbf{Linear} & \textbf{MLP} & \textbf{CosineCls} & \textbf{RouterDC} & \textbf{VLC} & \textbf{ZOOTER} \\
    \midrule
    \multirow{12}{*}{\rotatebox{90}{General}} & \multirow{2}{*}{HallusionBench} & Avg. Acc. $\uparrow$ & 99.46 & 71.89 & 57.84 & 40.54 & 43.24 & 66.49 & 70.27 & 68.65 & 68.65 & \underline{79.46} & \textbf{80.00} & 67.03 & 71.89 \\
     &  & Avg. Cost $\downarrow$ & 0.23 & 2.97 & 0.14 & \textbf{0.18} & \underline{0.21} & 1.67 & 2.89 & 2.17 & 3.02 & 4.18 & 2.82 & 3.03 & 2.16 \\
    \cmidrule{2-16}
     & \multirow{2}{*}{InfoVQA} & Avg. Acc. $\uparrow$ & 96.49 & 83.28 & 53.68 & 58.36 & 61.87 & \underline{81.10} & 80.60 & 77.93 & 81.10 & 80.60 & 81.10 & \textbf{82.61} & 80.77 \\
     &  & Avg. Cost $\downarrow$ & 0.37 & 2.19 & 0.14 & \textbf{0.24} & \underline{0.29} & 1.49 & 2.09 & 1.66 & 1.73 & 0.99 & 1.12 & 1.20 & 1.61 \\
    \cmidrule{2-16}
     & \multirow{2}{*}{MMBench} & Avg. Acc. $\uparrow$ & 98.16 & 89.86 & 69.59 & 69.59 & 81.11 & 85.25 & 70.97 & 82.95 & 87.56 & \textbf{90.32} & \underline{90.32} & 87.10 & 84.79 \\
     &  & Avg. Cost $\downarrow$ & 0.24 & 2.22 & 0.13 & \textbf{0.19} & \underline{0.27} & 2.26 & 1.17 & 1.55 & 1.90 & 5.33 & 2.52 & 1.84 & 1.57 \\
    \cmidrule{2-16}
     & \multirow{2}{*}{MMMU} & Avg. Acc. $\uparrow$ & 98.46 & 74.36 & 45.64 & 42.05 & 53.33 & 60.00 & 65.64 & 67.69 & \underline{76.92} & \textbf{77.44} & 76.41 & 75.38 & 72.31 \\
     &  & Avg. Cost $\downarrow$ & 0.64 & 3.24 & 0.17 & \textbf{0.41} & \underline{0.43} & 6.95 & 10.59 & 11.33 & 13.22 & 17.54 & 7.06 & 12.91 & 10.65 \\
    \cmidrule{2-16}
     & \multirow{2}{*}{MMStar} & Avg. Acc. $\uparrow$ & 96.68 & 72.76 & 49.17 & 54.48 & 59.80 & 68.44 & 66.11 & 63.46 & 70.76 & \textbf{75.08} & 72.76 & 69.44 & \underline{74.75} \\
     &  & Avg. Cost $\downarrow$ & 0.36 & 2.29 & 0.14 & \textbf{0.25} & \underline{0.29} & 2.85 & 1.68 & 1.76 & 2.78 & 7.98 & 8.26 & 2.58 & 2.78 \\
    \cmidrule{2-16}
     & \multirow{2}{*}{RealWorldQA} & Avg. Acc. $\uparrow$ & 99.32 & 77.40 & 67.12 & 55.48 & 63.70 & 67.81 & 60.27 & 69.86 & 69.86 & \underline{78.08} & 77.40 & 74.66 & \textbf{80.14} \\
     &  & Avg. Cost $\downarrow$ & 0.22 & 2.20 & 0.13 & \textbf{0.14} & \underline{0.18} & 1.46 & 1.17 & 1.15 & 1.84 & 3.55 & 3.77 & 2.27 & 1.78 \\
     \midrule
    \multirow{8}{*}{\rotatebox{90}{STEM}} & \multirow{2}{*}{AI2D} & Avg. Acc. $\uparrow$ & 99.69 & 88.21 & 72.33 & 67.45 & 73.58 & 80.66 & 76.42 & 76.10 & 83.80 & \textbf{87.11} & \underline{86.95} & 82.70 & 84.43 \\
    &  & Avg. Cost $\downarrow$ & 0.24 & 2.19 & 0.13 & \textbf{0.16} & \underline{0.19} & 2.05 & 2.56 & 1.20 & 2.94 & 7.17 & 2.10 & 2.02 & 2.40 \\
   \cmidrule{2-16}
    & \multirow{2}{*}{MathVerse} & Avg. Acc. $\uparrow$ & 90.36 & 49.74 & 23.05 & 59.77 & 67.97 & 68.75 & 66.93 & 69.53 & \textbf{71.09} & 62.37 & 66.15 & \underline{69.92} & 52.99 \\
    &  & Avg. Cost $\downarrow$ & 1.06 & 4.67 & 0.18 & \textbf{1.64} & \underline{1.70} & 4.01 & 3.19 & 3.99 & 4.04 & 7.85 & 2.59 & 3.87 & 2.55 \\
   \cmidrule{2-16}
    & \multirow{2}{*}{MathVision} & Avg. Acc. $\uparrow$ & 89.55 & 38.81 & 13.43 & 28.36 & 23.88 & 35.82 & 46.27 & 49.25 & 50.75 & \textbf{61.19} & \underline{61.19} & 50.75 & 61.19 \\
    &  & Avg. Cost $\downarrow$ & 1.77 & 8.16 & 0.17 & \underline{1.23} & \textbf{0.72} & 13.10 & 5.64 & 9.92 & 15.50 & 26.73 & 2.73 & 19.30 & 24.39 \\
   \cmidrule{2-16}
    & \multirow{2}{*}{MathVista} & Avg. Acc. $\uparrow$ & 95.85 & 75.13 & 51.30 & 53.89 & 70.98 & 77.20 & 73.58 & 72.54 & \underline{79.27} & 76.68 & 77.20 & 76.17 & \textbf{79.79} \\
    &  & Avg. Cost $\downarrow$ & 0.36 & 5.00 & 0.16 & \textbf{0.50} & \underline{0.54} & 3.46 & 2.81 & 3.79 & 4.82 & 9.52 & 6.67 & 3.54 & 4.45 \\
   \midrule
   \multirow{8}{*}{\rotatebox{90}{Chart \& Doc}} & \multirow{2}{*}{ChartQA} & Avg. Acc. $\uparrow$ & 97.80 & 86.20 & 81.00 & 81.40 & 80.80 & 81.00 & 40.60 & 70.20 & \textbf{83.80} & 70.60 & 76.40 & \underline{83.20} & 82.60 \\
    &  & Avg. Cost $\downarrow$ & 0.21 & 2.19 & 0.14 & \textbf{0.14} & \underline{0.17} & 1.09 & 0.69 & 0.38 & 0.98 & 1.74 & 0.86 & 1.30 & 1.34 \\
   \cmidrule{2-16}
    & \multirow{2}{*}{DocVQA} & Avg. Acc. $\uparrow$ & 98.56 & 91.48 & 78.95 & 78.66 & 80.48 & 85.55 & 82.30 & 80.19 & \underline{88.13} & 87.85 & 88.04 & 87.37 & \textbf{89.57} \\
    &  & Avg. Cost $\downarrow$ & 0.18 & 2.18 & 0.14 & \textbf{0.15} & \underline{0.18} & 1.22 & 1.25 & 0.58 & 1.34 & 0.98 & 0.21 & 1.34 & 1.41 \\
   \cmidrule{2-16}
    & \multirow{2}{*}{OCRBench} & Avg. Acc. $\uparrow$ & 99.06 & 89.20 & 80.28 & 77.00 & 81.22 & 85.92 & 79.34 & 81.22 & 86.85 & \underline{93.90} & \textbf{94.37} & 89.67 & 87.32 \\
    &  & Avg. Cost $\downarrow$ & 0.23 & 2.41 & 0.14 & \textbf{0.16} & \underline{0.19} & 1.44 & 0.96 & 0.71 & 1.61 & 1.60 & 0.68 & 1.77 & 1.70 \\
   \cmidrule{2-16}
    & \multirow{2}{*}{TextVQA} & Avg. Acc. $\uparrow$ & 89.46 & 73.85 & 72.13 & 70.98 & \underline{72.13} & 67.05 & 66.09 & 68.29 & 71.26 & 67.62 & 68.49 & \textbf{73.56} & 69.73 \\
    &  & Avg. Cost $\downarrow$ & 0.18 & 2.15 & 0.13 & \textbf{0.14} & \underline{0.16} & 1.12 & 1.81 & 0.33 & 1.86 & 1.31 & 0.84 & 1.41 & 0.78 \\
    \midrule
    \multirow{3}{*}{\rotatebox{90}{Average}} & \multirow{3}{*}{All Datasets} & Avg. Acc. $\uparrow$ & 95.60 & 78.01 & 62.43 & 66.26 & 70.68 & 75.49 & 70.01 & 73.08 & \underline{78.62} & 77.19 & 78.24 & \textbf{78.65} & 76.70 \\
    &  & Avg. Cost $\downarrow$ & 0.37 & 2.72 & 0.14 & \textbf{0.38} & \underline{0.41} & 2.18 & 2.21 & 1.82 & 2.75 & 3.37 & 1.14 & 2.53 & 2.32 \\
    &  & Rank Score $\uparrow$ & 93.68 & 68.88 & 64.63 & 67.13 & \underline{71.09} & 68.92 & 64.62 & 68.23 & 69.19 & 65.90 & \textbf{74.81} & 70.04 & 69.36 \\
    \bottomrule
    \end{tabular}
    }
\end{table*}
\newpage 
\begin{table*}[tbp]
    \centering
    \small
    \setlength{\tabcolsep}{3pt}
    \caption{Performance comparison of router methods on VL-RouterBench across datasets for $\lambda=10.0$. \textbf{Avg. Acc.} is average accuracy (\%), \textbf{Avg. Cost} is average cost (\$/10K samples). The best and second-best results among learned routers are highlighted in \textbf{bold} and \underline{underlined}, respectively (excluding Oracle and baseline methods).}
    \label{tab:dataset_results_arxiv_lambda_10.0}
    \vspace{-3pt}
    \resizebox{\linewidth}{!}{
        \begin{tabular}{c c c|c c c c c c c c c c c c c}
    \toprule
    \multirow{2}{*}{\textbf{Group}} & \multirow{2}{*}{\textbf{Dataset}} & \multirow{2}{*}{\textbf{Metric}} & \multicolumn{3}{c}{\textbf{Baselines}} & \multicolumn{4}{c}{\textbf{Hard Label}} & \multicolumn{6}{c}{\textbf{Soft Label ($\lambda=10.0$)}} \\
    \cmidrule(lr){4-6} \cmidrule(lr){7-10} \cmidrule(lr){11-16}
     &  &  & \textbf{Oracle} & \textbf{Strongest} & \textbf{Cheapest} & \textbf{KNN} & \textbf{PRkNN} & \textbf{OVR} & \textbf{K-means} & \textbf{Linear} & \textbf{MLP} & \textbf{CosineCls} & \textbf{RouterDC} & \textbf{VLC} & \textbf{ZOOTER} \\
    \midrule
    \multirow{12}{*}{\rotatebox{90}{General}} & \multirow{2}{*}{HallusionBench} & Avg. Acc. $\uparrow$ & 99.46 & 71.89 & 57.84 & 40.54 & 43.24 & 66.49 & 70.27 & \textbf{73.51} & 65.95 & 61.08 & 58.38 & \underline{70.81} & 65.41 \\
     &  & Avg. Cost $\downarrow$ & 0.23 & 2.97 & 0.14 & \underline{0.18} & 0.21 & 1.67 & 2.89 & 3.13 & 0.98 & \textbf{0.17} & 0.19 & 0.90 & 1.49 \\
    \cmidrule{2-16}
     & \multirow{2}{*}{InfoVQA} & Avg. Acc. $\uparrow$ & 96.49 & 83.28 & 53.68 & 58.36 & 61.87 & \underline{81.10} & 80.60 & 72.41 & 78.60 & 57.19 & 65.05 & \textbf{83.11} & 80.43 \\
     &  & Avg. Cost $\downarrow$ & 0.37 & 2.19 & 0.14 & \textbf{0.24} & 0.29 & 1.49 & 2.09 & 0.87 & 1.31 & \underline{0.24} & 0.43 & 2.00 & 1.72 \\
    \cmidrule{2-16}
     & \multirow{2}{*}{MMBench} & Avg. Acc. $\uparrow$ & 98.16 & 89.86 & 69.59 & 69.59 & 81.11 & 85.25 & 70.97 & 82.95 & \textbf{89.86} & 77.42 & 79.72 & 88.02 & \underline{89.86} \\
     &  & Avg. Cost $\downarrow$ & 0.24 & 2.22 & 0.13 & \textbf{0.19} & 0.27 & 2.26 & 1.17 & 2.44 & 1.08 & \underline{0.24} & 0.28 & 0.93 & 1.62 \\
    \cmidrule{2-16}
     & \multirow{2}{*}{MMMU} & Avg. Acc. $\uparrow$ & 98.46 & 74.36 & 45.64 & 42.05 & 53.33 & 60.00 & \underline{65.64} & \textbf{70.77} & 58.46 & 48.21 & 51.79 & 62.05 & 60.00 \\
     &  & Avg. Cost $\downarrow$ & 0.64 & 3.24 & 0.17 & 0.41 & 0.43 & 6.95 & 10.59 & 11.64 & 1.59 & \underline{0.33} & \textbf{0.31} & 1.08 & 1.16 \\
    \cmidrule{2-16}
     & \multirow{2}{*}{MMStar} & Avg. Acc. $\uparrow$ & 96.68 & 72.76 & 49.17 & 54.48 & 59.80 & 68.44 & 66.11 & 67.44 & 69.10 & 56.81 & 59.14 & \underline{69.77} & \textbf{70.43} \\
     &  & Avg. Cost $\downarrow$ & 0.36 & 2.29 & 0.14 & 0.25 & 0.29 & 2.85 & 1.68 & 2.33 & 1.05 & \textbf{0.24} & \underline{0.24} & 0.83 & 1.33 \\
    \cmidrule{2-16}
     & \multirow{2}{*}{RealWorldQA} & Avg. Acc. $\uparrow$ & 99.32 & 77.40 & 67.12 & 55.48 & 63.70 & 67.81 & 60.27 & 67.12 & 69.86 & 63.70 & 61.64 & \underline{71.92} & \textbf{79.45} \\
     &  & Avg. Cost $\downarrow$ & 0.22 & 2.20 & 0.13 & \textbf{0.14} & 0.18 & 1.46 & 1.17 & 1.91 & 0.70 & \underline{0.16} & 0.31 & 1.27 & 1.29 \\
     \midrule
    \multirow{8}{*}{\rotatebox{90}{STEM}} & \multirow{2}{*}{AI2D} & Avg. Acc. $\uparrow$ & 99.69 & 88.21 & 72.33 & 67.45 & 73.58 & 80.66 & 76.42 & 79.56 & 83.02 & 73.58 & 72.80 & \textbf{86.16} & \underline{84.28} \\
     &  & Avg. Cost $\downarrow$ & 0.24 & 2.19 & 0.13 & \textbf{0.16} & 0.19 & 2.05 & 2.56 & 2.20 & 1.10 & \underline{0.17} & 0.21 & 1.50 & 1.37 \\
    \cmidrule{2-16}
     & \multirow{2}{*}{MathVerse} & Avg. Acc. $\uparrow$ & 90.36 & 49.74 & 23.05 & 59.77 & 67.97 & \underline{68.75} & 66.93 & 67.71 & \textbf{69.92} & 64.71 & 57.68 & 62.11 & 64.32 \\
     &  & Avg. Cost $\downarrow$ & 1.06 & 4.67 & 0.18 & \underline{1.64} & 1.70 & 4.01 & 3.19 & 3.33 & 3.28 & 3.92 & 2.39 & \textbf{1.52} & 2.18 \\
    \cmidrule{2-16}
     & \multirow{2}{*}{MathVision} & Avg. Acc. $\uparrow$ & 89.55 & 38.81 & 13.43 & 28.36 & 23.88 & 35.82 & 46.27 & 46.27 & 49.25 & 49.25 & 37.31 & \textbf{52.24} & \underline{52.24} \\
     &  & Avg. Cost $\downarrow$ & 1.77 & 8.16 & 0.17 & 1.23 & \textbf{0.72} & 13.10 & 5.64 & 7.07 & 2.55 & 9.16 & \underline{1.17} & 1.78 & 1.78 \\
    \cmidrule{2-16}
     & \multirow{2}{*}{MathVista} & Avg. Acc. $\uparrow$ & 95.85 & 75.13 & 51.30 & 53.89 & 70.98 & 77.20 & 73.58 & 76.17 & \textbf{81.87} & 62.69 & 62.18 & \underline{77.72} & 77.72 \\
     &  & Avg. Cost $\downarrow$ & 0.36 & 5.00 & 0.16 & \underline{0.50} & 0.54 & 3.46 & 2.81 & 3.72 & 1.90 & 0.77 & \textbf{0.44} & 1.14 & 1.17 \\
     \midrule
    \multirow{8}{*}{\rotatebox{90}{Chart \& Doc}} & \multirow{2}{*}{ChartQA} & Avg. Acc. $\uparrow$ & 97.80 & 86.20 & 81.00 & 81.40 & 80.80 & 81.00 & 40.60 & 79.40 & \underline{84.00} & 82.40 & 83.40 & \textbf{87.00} & 81.20 \\
     &  & Avg. Cost $\downarrow$ & 0.21 & 2.19 & 0.14 & \textbf{0.14} & \underline{0.17} & 1.09 & 0.69 & 1.36 & 0.58 & 0.17 & 0.22 & 0.85 & 1.58 \\
    \cmidrule{2-16}
     & \multirow{2}{*}{DocVQA} & Avg. Acc. $\uparrow$ & 98.56 & 91.48 & 78.95 & 78.66 & 80.48 & 85.55 & 82.30 & 82.78 & 85.07 & 79.14 & 81.24 & \textbf{90.14} & \underline{88.52} \\
     &  & Avg. Cost $\downarrow$ & 0.18 & 2.18 & 0.14 & \textbf{0.15} & 0.18 & 1.22 & 1.25 & 0.62 & 0.65 & \underline{0.15} & 0.21 & 1.59 & 1.20 \\
    \cmidrule{2-16}
     & \multirow{2}{*}{OCRBench} & Avg. Acc. $\uparrow$ & 99.06 & 89.20 & 80.28 & 77.00 & 81.22 & 85.92 & 79.34 & 80.28 & \textbf{88.26} & 81.22 & 85.45 & \underline{88.26} & 85.45 \\
     &  & Avg. Cost $\downarrow$ & 0.23 & 2.41 & 0.14 & \textbf{0.16} & 0.19 & 1.44 & 0.96 & 1.18 & 1.06 & \underline{0.16} & 0.21 & 0.57 & 0.97 \\
    \cmidrule{2-16}
     & \multirow{2}{*}{TextVQA} & Avg. Acc. $\uparrow$ & 89.46 & 73.85 & 72.13 & 70.98 & 72.13 & 67.05 & 66.09 & 66.38 & \textbf{72.70} & \underline{72.41} & 70.59 & 70.88 & 71.74 \\
     &  & Avg. Cost $\downarrow$ & 0.18 & 2.15 & 0.13 & \textbf{0.14} & 0.16 & 1.12 & 1.81 & 1.05 & 0.57 & \underline{0.14} & 0.17 & 0.62 & 0.31 \\
    \midrule
    \multirow{3}{*}{\rotatebox{90}{Average}} & \multirow{3}{*}{All Datasets} & Avg. Acc. $\uparrow$ & 95.60 & 78.01 & 62.43 & 66.26 & 70.68 & 75.49 & 70.01 & 73.97 & \underline{77.32} & 69.88 & 70.99 & \textbf{78.09} & 77.26 \\
    &  & Avg. Cost $\downarrow$ & 0.37 & 2.72 & 0.14 & \textbf{0.38} & \underline{0.41} & 2.18 & 2.21 & 2.08 & 1.22 & 0.77 & 0.53 & 1.23 & 1.30 \\
    &  & Rank Score $\uparrow$ & 93.68 & 68.88 & 64.63 & 67.13 & 71.09 & 68.92 & 64.62 & 68.10 & \underline{73.73} & 68.98 & 70.89 & \textbf{74.33} & 73.40 \\
    \bottomrule
    \end{tabular}
    }
\end{table*}
\newpage 
\begin{table*}[tbp]
    \centering
    \small
    \setlength{\tabcolsep}{3pt}
    \caption{Performance comparison of router methods on VL-RouterBench across datasets for $\lambda=100.0$. \textbf{Avg. Acc.} is average accuracy (\%), \textbf{Avg. Cost} is average cost (\$/10K samples). The best and second-best results among learned routers are highlighted in \textbf{bold} and \underline{underlined}, respectively (excluding Oracle and baseline methods).}
    \label{tab:dataset_results_arxiv_lambda_100.0}
    \vspace{-3pt}
    \resizebox{\linewidth}{!}{
        \begin{tabular}{c c c|c c c c c c c c c c c c c}
    \toprule
    \multirow{2}{*}{\textbf{Group}} & \multirow{2}{*}{\textbf{Dataset}} & \multirow{2}{*}{\textbf{Metric}} & \multicolumn{3}{c}{\textbf{Baselines}} & \multicolumn{4}{c}{\textbf{Hard Label}} & \multicolumn{6}{c}{\textbf{Soft Label ($\lambda=100.0$)}} \\
    \cmidrule(lr){4-6} \cmidrule(lr){7-10} \cmidrule(lr){11-16}
     &  &  & \textbf{Oracle} & \textbf{Strongest} & \textbf{Cheapest} & \textbf{KNN} & \textbf{PRkNN} & \textbf{OVR} & \textbf{K-means} & \textbf{Linear} & \textbf{MLP} & \textbf{CosineCls} & \textbf{RouterDC} & \textbf{VLC} & \textbf{ZOOTER} \\
    \midrule
    \multirow{12}{*}{\rotatebox{90}{General}} & \multirow{2}{*}{HallusionBench} & Avg. Acc. $\uparrow$ & 99.46 & 71.89 & 57.84 & 40.54 & 43.24 & 66.49 & \textbf{70.27} & \underline{67.03} & 58.92 & 56.22 & 59.46 & 59.46 & 58.38 \\
     &  & Avg. Cost $\downarrow$ & 0.23 & 2.97 & 0.14 & \textbf{0.18} & 0.21 & 1.67 & 2.89 & 1.56 & 0.33 & 0.20 & 0.21 & 0.23 & \underline{0.19} \\
    \cmidrule{2-16}
     & \multirow{2}{*}{InfoVQA} & Avg. Acc. $\uparrow$ & 96.49 & 83.28 & 53.68 & 58.36 & 61.87 & \textbf{81.10} & \underline{80.60} & 79.77 & 74.41 & 65.05 & 68.06 & 67.56 & 56.69 \\
     &  & Avg. Cost $\downarrow$ & 0.37 & 2.19 & 0.14 & \underline{0.24} & 0.29 & 1.49 & 2.09 & 1.71 & 0.56 & 0.36 & 0.44 & 0.60 & \textbf{0.21} \\
    \cmidrule{2-16}
     & \multirow{2}{*}{MMBench} & Avg. Acc. $\uparrow$ & 98.16 & 89.86 & 69.59 & 69.59 & 81.11 & \textbf{85.25} & 70.97 & \underline{83.41} & 77.42 & 80.18 & 81.57 & 82.49 & 68.66 \\
     &  & Avg. Cost $\downarrow$ & 0.24 & 2.22 & 0.13 & \textbf{0.19} & 0.27 & 2.26 & 1.17 & 1.32 & 0.30 & 0.23 & 0.44 & 0.34 & \underline{0.20} \\
    \cmidrule{2-16}
     & \multirow{2}{*}{MMMU} & Avg. Acc. $\uparrow$ & 98.46 & 74.36 & 45.64 & 42.05 & 53.33 & 60.00 & \underline{65.64} & \textbf{68.72} & 55.90 & 51.28 & 53.85 & 53.33 & 54.87 \\
     &  & Avg. Cost $\downarrow$ & 0.64 & 3.24 & 0.17 & \textbf{0.41} & 0.43 & 6.95 & 10.59 & 10.21 & 0.52 & \underline{0.41} & 0.45 & 0.50 & 0.49 \\
    \cmidrule{2-16}
     & \multirow{2}{*}{MMStar} & Avg. Acc. $\uparrow$ & 96.68 & 72.76 & 49.17 & 54.48 & 59.80 & \textbf{68.44} & \underline{66.11} & 62.13 & 64.45 & 60.47 & 58.80 & 60.47 & 61.13 \\
     &  & Avg. Cost $\downarrow$ & 0.36 & 2.29 & 0.14 & \textbf{0.25} & \underline{0.29} & 2.85 & 1.68 & 2.66 & 0.36 & 0.30 & 0.30 & 0.32 & 0.30 \\
    \cmidrule{2-16}
     & \multirow{2}{*}{RealWorldQA} & Avg. Acc. $\uparrow$ & 99.32 & 77.40 & 67.12 & 55.48 & 63.70 & \underline{67.81} & 60.27 & 65.07 & 67.12 & 65.07 & 65.75 & \textbf{70.55} & 66.44 \\
     &  & Avg. Cost $\downarrow$ & 0.22 & 2.20 & 0.13 & \textbf{0.14} & 0.18 & 1.46 & 1.17 & 0.80 & 0.17 & 0.23 & 0.21 & 0.27 & \underline{0.15} \\
     \midrule
    \multirow{8}{*}{\rotatebox{90}{STEM}} & \multirow{2}{*}{AI2D} & Avg. Acc. $\uparrow$ & 99.69 & 88.21 & 72.33 & 67.45 & 73.58 & \underline{80.66} & 76.42 & 78.14 & \textbf{80.82} & 74.53 & 72.01 & 74.53 & 72.96 \\
     &  & Avg. Cost $\downarrow$ & 0.24 & 2.19 & 0.13 & \textbf{0.16} & \underline{0.19} & 2.05 & 2.56 & 0.91 & 0.36 & 0.25 & 0.21 & 0.22 & 0.23 \\
    \cmidrule{2-16}
     & \multirow{2}{*}{MathVerse} & Avg. Acc. $\uparrow$ & 90.36 & 49.74 & 23.05 & 59.77 & \underline{67.97} & \textbf{68.75} & 66.93 & 67.32 & 61.98 & 64.84 & 64.06 & 60.29 & 63.02 \\
     &  & Avg. Cost $\downarrow$ & 1.06 & 4.67 & 0.18 & 1.64 & 1.70 & 4.01 & 3.19 & 3.55 & \underline{1.60} & 3.86 & 3.69 & \textbf{1.37} & 2.31 \\
    \cmidrule{2-16}
     & \multirow{2}{*}{MathVision} & Avg. Acc. $\uparrow$ & 89.55 & 38.81 & 13.43 & 28.36 & 23.88 & 35.82 & 46.27 & 41.79 & 49.25 & \textbf{52.24} & 50.75 & 50.75 & \underline{52.24} \\
     &  & Avg. Cost $\downarrow$ & 1.77 & 8.16 & 0.17 & \underline{1.23} & \textbf{0.72} & 13.10 & 5.64 & 10.53 & 1.80 & 3.29 & 2.49 & 1.76 & 1.78 \\
    \cmidrule{2-16}
     & \multirow{2}{*}{MathVista} & Avg. Acc. $\uparrow$ & 95.85 & 75.13 & 51.30 & 53.89 & 70.98 & \underline{77.20} & 73.58 & 71.50 & 75.65 & 65.29 & 68.39 & 73.06 & \textbf{77.72} \\
     &  & Avg. Cost $\downarrow$ & 0.36 & 5.00 & 0.16 & \textbf{0.50} & \underline{0.54} & 3.46 & 2.81 & 4.93 & 1.03 & 0.85 & 0.79 & 0.89 & 1.11 \\
     \midrule
    \multirow{8}{*}{\rotatebox{90}{Chart \& Doc}} & \multirow{2}{*}{ChartQA} & Avg. Acc. $\uparrow$ & 97.80 & 86.20 & 81.00 & 81.40 & 80.80 & 81.00 & 40.60 & 80.60 & 81.40 & \underline{83.00} & 82.00 & \textbf{83.80} & 81.80 \\
     &  & Avg. Cost $\downarrow$ & 0.21 & 2.19 & 0.14 & \textbf{0.14} & 0.17 & 1.09 & 0.69 & 0.91 & \underline{0.16} & 0.21 & 0.25 & 0.25 & 0.16 \\
    \cmidrule{2-16}
     & \multirow{2}{*}{DocVQA} & Avg. Acc. $\uparrow$ & 98.56 & 91.48 & 78.95 & 78.66 & 80.48 & \textbf{85.55} & 82.30 & 81.44 & \underline{83.06} & 80.48 & 79.52 & 82.58 & 79.04 \\
     &  & Avg. Cost $\downarrow$ & 0.18 & 2.18 & 0.14 & \textbf{0.15} & 0.18 & 1.22 & 1.25 & 0.74 & 0.36 & 0.20 & 0.21 & 0.29 & \underline{0.16} \\
    \cmidrule{2-16}
     & \multirow{2}{*}{OCRBench} & Avg. Acc. $\uparrow$ & 99.06 & 89.20 & 80.28 & 77.00 & 81.22 & \underline{85.92} & 79.34 & 78.40 & 83.10 & 84.04 & 83.57 & \textbf{87.32} & 79.34 \\
     &  & Avg. Cost $\downarrow$ & 0.23 & 2.41 & 0.14 & \textbf{0.16} & 0.19 & 1.44 & 0.96 & 0.78 & 0.28 & 0.20 & 0.22 & 0.32 & \underline{0.17} \\
    \cmidrule{2-16}
     & \multirow{2}{*}{TextVQA} & Avg. Acc. $\uparrow$ & 89.46 & 73.85 & 72.13 & 70.98 & \textbf{72.13} & 67.05 & 66.09 & 64.37 & 70.31 & 71.46 & 70.31 & 70.59 & \underline{72.13} \\
     &  & Avg. Cost $\downarrow$ & 0.18 & 2.15 & 0.13 & \underline{0.14} & 0.16 & 1.12 & 1.81 & 0.39 & 0.20 & 0.17 & 0.15 & 0.25 & \textbf{0.13} \\
    \midrule
    \multirow{3}{*}{\rotatebox{90}{Average}} & \multirow{3}{*}{All Datasets} & Avg. Acc. $\uparrow$ & 95.60 & 78.01 & 62.43 & 66.26 & 70.68 & \textbf{75.49} & 70.01 & 73.20 & \underline{73.31} & 71.35 & 72.07 & 72.02 & 69.97 \\
    &  & Avg. Cost $\downarrow$ & 0.37 & 2.72 & 0.14 & \textbf{0.38} & \underline{0.41} & 2.18 & 2.21 & 1.85 & 0.52 & 0.74 & 0.73 & 0.48 & 0.50 \\
    &  & Rank Score $\uparrow$ & 93.68 & 68.88 & 64.63 & 67.13 & 71.09 & 68.92 & 64.62 & 68.26 & \textbf{73.01} & 70.38 & 71.07 & \underline{72.01} & 70.06 \\
    \bottomrule
    \end{tabular}
    }
\end{table*}
\newpage 
\begin{table*}[tbp]
    \centering
    \small
    \setlength{\tabcolsep}{3pt}
    \caption{Performance comparison of router methods on VL-RouterBench across datasets for $\lambda=1000.0$. \textbf{Avg. Acc.} is average accuracy (\%), \textbf{Avg. Cost} is average cost (\$/10K samples). The best and second-best results among learned routers are highlighted in \textbf{bold} and \underline{underlined}, respectively (excluding Oracle and baseline methods).}
    \label{tab:dataset_results_arxiv_lambda_1000.0}
    \vspace{-3pt}
    \resizebox{\linewidth}{!}{
        \begin{tabular}{c c c|c c c c c c c c c c c c c}
    \toprule
    \multirow{2}{*}{\textbf{Group}} & \multirow{2}{*}{\textbf{Dataset}} & \multirow{2}{*}{\textbf{Metric}} & \multicolumn{3}{c}{\textbf{Baselines}} & \multicolumn{4}{c}{\textbf{Hard Label}} & \multicolumn{6}{c}{\textbf{Soft Label ($\lambda=1000.0$)}} \\
    \cmidrule(lr){4-6} \cmidrule(lr){7-10} \cmidrule(lr){11-16}
     &  &  & \textbf{Oracle} & \textbf{Strongest} & \textbf{Cheapest} & \textbf{KNN} & \textbf{PRkNN} & \textbf{OVR} & \textbf{K-means} & \textbf{Linear} & \textbf{MLP} & \textbf{CosineCls} & \textbf{RouterDC} & \textbf{VLC} & \textbf{ZOOTER} \\
    \midrule
    \multirow{12}{*}{\rotatebox{90}{General}} & \multirow{2}{*}{HallusionBench} & Avg. Acc. $\uparrow$ & 99.46 & 71.89 & 57.84 & 40.54 & 43.24 & 66.49 & \textbf{70.27} & \underline{68.11} & 46.49 & 57.84 & 56.22 & 57.30 & 57.30 \\
     &  & Avg. Cost $\downarrow$ & 0.23 & 2.97 & 0.14 & 0.18 & 0.21 & 1.67 & 2.89 & 1.62 & 0.18 & \textbf{0.14} & 0.20 & \underline{0.14} & 0.14 \\
    \cmidrule{2-16}
     & \multirow{2}{*}{InfoVQA} & Avg. Acc. $\uparrow$ & 96.49 & 83.28 & 53.68 & 58.36 & 61.87 & \underline{81.10} & 80.60 & \textbf{81.27} & 59.36 & 55.18 & 65.05 & 54.01 & 53.68 \\
     &  & Avg. Cost $\downarrow$ & 0.37 & 2.19 & 0.14 & 0.24 & 0.29 & 1.49 & 2.09 & 1.17 & 0.22 & 0.16 & 0.45 & \underline{0.15} & \textbf{0.14} \\
    \cmidrule{2-16}
     & \multirow{2}{*}{MMBench} & Avg. Acc. $\uparrow$ & 98.16 & 89.86 & 69.59 & 69.59 & \underline{81.11} & \textbf{85.25} & 70.97 & 80.18 & 76.96 & 73.27 & 77.88 & 72.35 & 70.97 \\
     &  & Avg. Cost $\downarrow$ & 0.24 & 2.22 & 0.13 & 0.19 & 0.27 & 2.26 & 1.17 & 1.43 & 0.21 & 0.19 & 0.23 & \underline{0.17} & \textbf{0.13} \\
    \cmidrule{2-16}
     & \multirow{2}{*}{MMMU} & Avg. Acc. $\uparrow$ & 98.46 & 74.36 & 45.64 & 42.05 & 53.33 & 60.00 & \textbf{65.64} & \underline{61.03} & 47.69 & 49.74 & 54.36 & 45.13 & 43.08 \\
     &  & Avg. Cost $\downarrow$ & 0.64 & 3.24 & 0.17 & 0.41 & 0.43 & 6.95 & 10.59 & 4.69 & 0.26 & 0.33 & 0.52 & \underline{0.18} & \textbf{0.17} \\
    \cmidrule{2-16}
     & \multirow{2}{*}{MMStar} & Avg. Acc. $\uparrow$ & 96.68 & 72.76 & 49.17 & 54.48 & 59.80 & \textbf{68.44} & \underline{66.11} & 64.45 & 54.82 & 51.50 & 57.81 & 49.50 & 49.17 \\
     &  & Avg. Cost $\downarrow$ & 0.36 & 2.29 & 0.14 & 0.25 & 0.29 & 2.85 & 1.68 & 1.79 & 0.22 & 0.19 & 0.32 & \underline{0.15} & \textbf{0.14} \\
    \cmidrule{2-16}
     & \multirow{2}{*}{RealWorldQA} & Avg. Acc. $\uparrow$ & 99.32 & 77.40 & 67.12 & 55.48 & 63.70 & \textbf{67.81} & 60.27 & 65.75 & 61.64 & \underline{67.12} & 65.07 & 65.75 & 67.12 \\
     &  & Avg. Cost $\downarrow$ & 0.22 & 2.20 & 0.13 & 0.14 & 0.18 & 1.46 & 1.17 & 1.06 & 0.14 & 0.14 & 0.28 & \textbf{0.13} & \underline{0.13} \\
    \midrule
    \multirow{8}{*}{\rotatebox{90}{STEM}} & \multirow{2}{*}{AI2D} & Avg. Acc. $\uparrow$ & 99.69 & 88.21 & 72.33 & 67.45 & 73.58 & \textbf{80.66} & \underline{76.42} & 76.10 & 71.86 & 72.33 & 75.31 & 72.96 & 72.01 \\
     &  & Avg. Cost $\downarrow$ & 0.24 & 2.19 & 0.13 & 0.16 & 0.19 & 2.05 & 2.56 & 1.51 & 0.15 & 0.14 & 0.24 & \textbf{0.13} & \underline{0.13} \\
    \cmidrule{2-16}
     & \multirow{2}{*}{MathVerse} & Avg. Acc. $\uparrow$ & 90.36 & 49.74 & 23.05 & 59.77 & \underline{67.97} & \textbf{68.75} & 66.93 & 66.93 & 56.90 & 60.29 & 64.19 & 40.89 & 23.05 \\
     &  & Avg. Cost $\downarrow$ & 1.06 & 4.67 & 0.18 & 1.64 & 1.70 & 4.01 & 3.19 & 3.19 & 1.35 & 1.37 & 3.75 & \underline{0.85} & \textbf{0.18} \\
    \cmidrule{2-16}
     & \multirow{2}{*}{MathVision} & Avg. Acc. $\uparrow$ & 89.55 & 38.81 & 13.43 & 28.36 & 23.88 & 35.82 & 46.27 & \underline{47.76} & 31.34 & \textbf{52.24} & 43.28 & 31.34 & 13.43 \\
     &  & Avg. Cost $\downarrow$ & 1.77 & 8.16 & 0.17 & 1.23 & \underline{0.72} & 13.10 & 5.64 & 5.56 & 1.03 & 1.78 & 1.47 & 1.01 & \textbf{0.17} \\
    \cmidrule{2-16}
     & \multirow{2}{*}{MathVista} & Avg. Acc. $\uparrow$ & 95.85 & 75.13 & 51.30 & 53.89 & 70.98 & \textbf{77.20} & \underline{73.58} & 72.02 & 56.99 & 67.88 & 62.18 & 54.92 & 51.30 \\
     &  & Avg. Cost $\downarrow$ & 0.36 & 5.00 & 0.16 & 0.50 & 0.54 & 3.46 & 2.81 & 2.75 & 0.37 & 0.80 & 0.52 & \underline{0.28} & \textbf{0.16} \\
    \midrule
    \multirow{8}{*}{\rotatebox{90}{Chart \& Doc}} & \multirow{2}{*}{ChartQA} & Avg. Acc. $\uparrow$ & 97.80 & 86.20 & 81.00 & 81.40 & 80.80 & 81.00 & 40.60 & 80.20 & 81.40 & \underline{81.60} & \textbf{84.20} & 81.00 & 81.00 \\
     &  & Avg. Cost $\downarrow$ & 0.21 & 2.19 & 0.14 & \textbf{0.14} & 0.17 & 1.09 & 0.69 & 0.60 & 0.18 & 0.15 & 0.24 & \underline{0.14} & 0.14 \\
    \cmidrule{2-16}
     & \multirow{2}{*}{DocVQA} & Avg. Acc. $\uparrow$ & 98.56 & 91.48 & 78.95 & 78.66 & 80.48 & \textbf{85.55} & \underline{82.30} & 81.05 & 79.43 & 78.95 & 80.67 & 79.23 & 78.85 \\
     &  & Avg. Cost $\downarrow$ & 0.18 & 2.18 & 0.14 & 0.15 & 0.18 & 1.22 & 1.25 & 0.70 & 0.15 & \textbf{0.14} & 0.22 & \underline{0.14} & 0.14 \\
    \cmidrule{2-16}
     & \multirow{2}{*}{OCRBench} & Avg. Acc. $\uparrow$ & 99.06 & 89.20 & 80.28 & 77.00 & 81.22 & \textbf{85.92} & 79.34 & 80.75 & 80.75 & 80.75 & \underline{83.57} & 80.75 & 80.28 \\
     &  & Avg. Cost $\downarrow$ & 0.23 & 2.41 & 0.14 & 0.16 & 0.19 & 1.44 & 0.96 & 0.79 & 0.20 & 0.15 & 0.19 & \textbf{0.14} & \underline{0.14} \\
    \cmidrule{2-16}
     & \multirow{2}{*}{TextVQA} & Avg. Acc. $\uparrow$ & 89.46 & 73.85 & 72.13 & 70.98 & \underline{72.13} & 67.05 & 66.09 & 66.95 & 71.65 & 72.13 & 71.55 & \textbf{72.22} & 72.13 \\
     &  & Avg. Cost $\downarrow$ & 0.18 & 2.15 & 0.13 & 0.14 & 0.16 & 1.12 & 1.81 & 1.14 & 0.14 & \textbf{0.13} & 0.16 & \underline{0.13} & 0.13 \\
    \midrule
    \multirow{3}{*}{\rotatebox{90}{Average}} & \multirow{3}{*}{All Datasets} & Avg. Acc. $\uparrow$ & 95.60 & 78.01 & 62.43 & 66.26 & 70.68 & \textbf{75.49} & 70.01 & \underline{73.40} & 67.75 & 68.65 & 71.17 & 65.21 & 62.33 \\
    &  & Avg. Cost $\downarrow$ & 0.37 & 2.72 & 0.14 & 0.38 & 0.41 & 2.18 & 2.21 & 1.57 & 0.33 & 0.35 & 0.72 & \underline{0.24} & \textbf{0.14} \\
    &  & Rank Score $\uparrow$ & 93.68 & 68.88 & 64.63 & 67.13 & \textbf{71.09} & \underline{68.92} & 64.62 & 69.29 & 68.73 & 69.53 & \underline{70.31} & 66.78 & 64.54 \\
    \bottomrule
    \end{tabular}
    }
\end{table*}
\newpage 
\begin{table*}[tbp]
    \centering
    \small
    \setlength{\tabcolsep}{3pt}
    \caption{Performance comparison of router methods on VL-RouterBench across datasets for $\lambda=10000.0$. \textbf{Avg. Acc.} is average accuracy (\%), \textbf{Avg. Cost} is average cost (\$/10K samples). The best and second-best results among learned routers are highlighted in \textbf{bold} and \underline{underlined}, respectively (excluding Oracle and baseline methods).}
    \label{tab:dataset_results_arxiv_lambda_10000.0}
    \vspace{-3pt}
    \resizebox{\linewidth}{!}{
    \begin{tabular}{c c c|c c c c c c c c c c c c c}
    \toprule
    \multirow{2}{*}{\textbf{Group}} & \multirow{2}{*}{\textbf{Dataset}} & \multirow{2}{*}{\textbf{Metric}} & \multicolumn{3}{c}{\textbf{Baselines}} & \multicolumn{4}{c}{\textbf{Hard Label}} & \multicolumn{6}{c}{\textbf{Soft Label ($\lambda=10000.0$)}} \\
    \cmidrule(lr){4-6} \cmidrule(lr){7-10} \cmidrule(lr){11-16}
     &  &  & \textbf{Oracle} & \textbf{Strongest} & \textbf{Cheapest} & \textbf{KNN} & \textbf{PRkNN} & \textbf{OVR} & \textbf{K-means} & \textbf{Linear} & \textbf{MLP} & \textbf{CosineCls} & \textbf{RouterDC} & \textbf{VLC} & \textbf{ZOOTER} \\
    \midrule
    \multirow{12}{*}{\rotatebox{90}{General}} & \multirow{2}{*}{HallusionBench} & Avg. Acc. $\uparrow$ & 99.46 & 71.89 & 57.84 & 40.54 & 43.24 & \underline{66.49} & \textbf{70.27} & 62.70 & 48.11 & 51.89 & 58.38 & 41.62 & 56.76 \\
     &  & Avg. Cost $\downarrow$ & 0.23 & 2.97 & 0.14 & 0.18 & 0.21 & 1.67 & 2.89 & 0.77 & \textbf{0.14} & 0.30 & 0.17 & 0.16 & \underline{0.14} \\
    \cmidrule{2-16}
     & \multirow{2}{*}{InfoVQA} & Avg. Acc. $\uparrow$ & 96.49 & 83.28 & 53.68 & 58.36 & 61.87 & \textbf{81.10} & \underline{80.60} & 79.10 & 58.19 & 62.54 & 61.71 & 60.87 & 53.68 \\
     &  & Avg. Cost $\downarrow$ & 0.37 & 2.19 & 0.14 & 0.24 & 0.29 & 1.49 & 2.09 & 0.80 & \underline{0.20} & 0.46 & 0.33 & 0.29 & \textbf{0.14} \\
    \cmidrule{2-16}
     & \multirow{2}{*}{MMBench} & Avg. Acc. $\uparrow$ & 98.16 & 89.86 & 69.59 & 69.59 & 81.11 & \textbf{85.25} & 70.97 & 71.43 & 74.19 & 76.50 & \underline{82.03} & 77.42 & 69.59 \\
     &  & Avg. Cost $\downarrow$ & 0.24 & 2.22 & 0.13 & 0.19 & 0.27 & 2.26 & 1.17 & 0.43 & \underline{0.16} & 0.61 & 0.47 & 0.19 & \textbf{0.13} \\
    \cmidrule{2-16}
     & \multirow{2}{*}{MMMU} & Avg. Acc. $\uparrow$ & 98.46 & 74.36 & 45.64 & 42.05 & 53.33 & \underline{60.00} & \textbf{65.64} & 55.38 & 44.10 & 53.33 & 45.13 & 44.10 & 45.13 \\
     &  & Avg. Cost $\downarrow$ & 0.64 & 3.24 & 0.17 & 0.41 & 0.43 & 6.95 & 10.59 & 1.20 & \underline{0.28} & 8.07 & 6.11 & 0.30 & \textbf{0.17} \\
    \cmidrule{2-16}
     & \multirow{2}{*}{MMStar} & Avg. Acc. $\uparrow$ & 96.68 & 72.76 & 49.17 & 54.48 & 59.80 & \textbf{68.44} & \underline{66.11} & 51.83 & 53.82 & 60.13 & 55.81 & 54.82 & 49.17 \\
     &  & Avg. Cost $\downarrow$ & 0.36 & 2.29 & 0.14 & 0.25 & 0.29 & 2.85 & 1.68 & 0.49 & \underline{0.18} & 3.12 & 1.77 & 0.20 & \textbf{0.14} \\
    \cmidrule{2-16}
     & \multirow{2}{*}{RealWorldQA} & Avg. Acc. $\uparrow$ & 99.32 & 77.40 & 67.12 & 55.48 & 63.70 & \textbf{67.81} & 60.27 & 60.96 & 65.75 & 59.59 & 62.33 & 62.33 & \underline{67.12} \\
     &  & Avg. Cost $\downarrow$ & 0.22 & 2.20 & 0.13 & 0.14 & 0.18 & 1.46 & 1.17 & 0.46 & \textbf{0.13} & 0.51 & 0.19 & 0.16 & \underline{0.13} \\
    \midrule
    \multirow{8}{*}{\rotatebox{90}{STEM}} & \multirow{2}{*}{AI2D} & Avg. Acc. $\uparrow$ & 99.69 & 88.21 & 72.33 & 67.45 & 73.58 & \textbf{80.66} & \underline{76.42} & 70.44 & 70.91 & 73.58 & 73.43 & 72.17 & 72.33 \\
     &  & Avg. Cost $\downarrow$ & 0.24 & 2.19 & 0.13 & 0.16 & 0.19 & 2.05 & 2.56 & 0.51 & \underline{0.14} & 0.38 & 0.27 & 0.15 & \textbf{0.13} \\
    \cmidrule{2-16}
     & \multirow{2}{*}{MathVerse} & Avg. Acc. $\uparrow$ & 90.36 & 49.74 & 23.05 & 59.77 & \underline{67.97} & \textbf{68.75} & 66.93 & 53.39 & 52.47 & 45.83 & 50.39 & 43.36 & 23.05 \\
     &  & Avg. Cost $\downarrow$ & 1.06 & 4.67 & 0.18 & 1.64 & 1.70 & 4.01 & 3.19 & 1.56 & 1.73 & 9.42 & 7.30 & \underline{0.79} & \textbf{0.18} \\
    \cmidrule{2-16}
     & \multirow{2}{*}{MathVision} & Avg. Acc. $\uparrow$ & 89.55 & 38.81 & 13.43 & 28.36 & 23.88 & \underline{35.82} & \textbf{46.27} & 34.33 & 31.34 & 23.88 & 22.39 & 22.39 & 13.43 \\
     &  & Avg. Cost $\downarrow$ & 1.77 & 8.16 & 0.17 & 1.23 & \underline{0.72} & 13.10 & 5.64 & 1.48 & 1.48 & 40.17 & 40.70 & 0.98 & \textbf{0.17} \\
    \cmidrule{2-16}
     & \multirow{2}{*}{MathVista} & Avg. Acc. $\uparrow$ & 95.85 & 75.13 & 51.30 & 53.89 & 70.98 & \textbf{77.20} & \underline{73.58} & 63.21 & 55.44 & 55.96 & 59.07 & 54.92 & 51.30 \\
     &  & Avg. Cost $\downarrow$ & 0.36 & 5.00 & 0.16 & 0.50 & 0.54 & 3.46 & 2.81 & 1.00 & 0.38 & 6.74 & 3.64 & \underline{0.31} & \textbf{0.16} \\
    \midrule
    \multirow{8}{*}{\rotatebox{90}{Chart \& Doc}} & \multirow{2}{*}{ChartQA} & Avg. Acc. $\uparrow$ & 97.80 & 86.20 & 81.00 & \underline{81.40} & 80.80 & 81.00 & 40.60 & 76.60 & 81.40 & 81.40 & \textbf{83.00} & 81.20 & 81.00 \\
     &  & Avg. Cost $\downarrow$ & 0.21 & 2.19 & 0.14 & \textbf{0.14} & 0.17 & 1.09 & 0.69 & 0.33 & 0.15 & 0.24 & 0.22 & 0.17 & \underline{0.14} \\
    \cmidrule{2-16}
     & \multirow{2}{*}{DocVQA} & Avg. Acc. $\uparrow$ & 98.56 & 91.48 & 78.95 & 78.66 & 80.48 & \textbf{85.55} & \underline{82.30} & 80.10 & 78.56 & 79.90 & 79.71 & 78.95 & 78.85 \\
     &  & Avg. Cost $\downarrow$ & 0.18 & 2.18 & 0.14 & 0.15 & 0.18 & 1.22 & 1.25 & 0.61 & \textbf{0.14} & 0.19 & 0.18 & 0.16 & \underline{0.14} \\
    \cmidrule{2-16}
     & \multirow{2}{*}{OCRBench} & Avg. Acc. $\uparrow$ & 99.06 & 89.20 & 80.28 & 77.00 & 81.22 & \textbf{85.92} & 79.34 & 80.28 & 79.34 & 83.10 & \underline{83.57} & 81.69 & 80.28 \\
     &  & Avg. Cost $\downarrow$ & 0.23 & 2.41 & 0.14 & \underline{0.16} & 0.19 & 1.44 & 0.96 & 0.50 & 0.17 & 0.22 & 0.21 & 0.16 & \textbf{0.14} \\
    \cmidrule{2-16}
     & \multirow{2}{*}{TextVQA} & Avg. Acc. $\uparrow$ & 89.46 & 73.85 & 72.13 & 70.98 & \textbf{72.13} & 67.05 & 66.09 & 66.67 & 71.55 & 70.11 & 70.59 & 69.92 & \underline{72.13} \\
     &  & Avg. Cost $\downarrow$ & 0.18 & 2.15 & 0.13 & 0.14 & 0.16 & 1.12 & 1.81 & 0.43 & \textbf{0.13} & 0.27 & 0.17 & 0.14 & \underline{0.13} \\
    \midrule
    \multirow{3}{*}{\rotatebox{90}{Average}} & \multirow{3}{*}{All Datasets} & Avg. Acc. $\uparrow$ & 95.60 & 78.01 & 62.43 & 66.26 & \underline{70.68} & \textbf{75.49} & 70.01 & 68.55 & 66.60 & 76.17 & 76.91 & 65.47 & 62.36 \\
    &  & Avg. Cost $\downarrow$ & 0.37 & 2.72 & 0.14 & 0.38 & 0.41 & 2.18 & 2.21 & 0.71 & 0.38 & 2.48 & 1.93 & \underline{0.27} & \textbf{0.14} \\
    &  & Rank Score $\uparrow$ & 93.68 & 68.88 & 64.63 & 67.13 & \textbf{71.09} & \underline{68.92} & 64.62 & 68.01 & 67.48 & 68.41 & 70.86 & 66.90 & 64.57 \\
    \bottomrule
    \end{tabular}
    }
\end{table*}
\newpage 
\begin{table*}[tbp]
    \centering
    \small
    \setlength{\tabcolsep}{3pt}
    \caption{Performance comparison of router methods on VL-RouterBench across datasets for $\lambda=\infty$. \textbf{Avg. Acc.} is average accuracy (\%), \textbf{Avg. Cost} is average cost (\$/10K samples). The best and second-best results among learned routers are highlighted in \textbf{bold} and \underline{underlined}, respectively (excluding Oracle and baseline methods).}
    \label{tab:dataset_results_arxiv_lambda_inf}
    \vspace{-3pt}
    \resizebox{\linewidth}{!}{
    \begin{tabular}{c c c|c c c c c c c c c c c c c}
    \toprule
    \multirow{2}{*}{\textbf{Group}} & \multirow{2}{*}{\textbf{Dataset}} & \multirow{2}{*}{\textbf{Metric}} & \multicolumn{3}{c}{\textbf{Baselines}} & \multicolumn{4}{c}{\textbf{Hard Label}} & \multicolumn{6}{c}{\textbf{Soft Label ($\lambda=\infty$)}} \\
    \cmidrule(lr){4-6} \cmidrule(lr){7-10} \cmidrule(lr){11-16}
     &  &  & \textbf{Oracle} & \textbf{Strongest} & \textbf{Cheapest} & \textbf{KNN} & \textbf{PRkNN} & \textbf{OVR} & \textbf{K-means} & \textbf{Linear} & \textbf{MLP} & \textbf{CosineCls} & \textbf{RouterDC} & \textbf{VLC} & \textbf{ZOOTER} \\
    \midrule
    \multirow{12}{*}{\rotatebox{90}{General}} & \multirow{2}{*}{HallusionBench} & Avg. Acc. $\uparrow$ & 99.46 & 71.89 & 57.84 & 40.54 & 43.24 & 66.49 & 70.27 & 57.84 & 47.57 & \underline{78.38} & \textbf{80.00} & 45.41 & 49.73 \\
     &  & Avg. Cost $\downarrow$ & 0.23 & 2.97 & 0.14 & 0.18 & 0.21 & 1.67 & 2.89 & \textbf{0.14} & 0.16 & 5.18 & 4.66 & 0.15 & \underline{0.14} \\
    \cmidrule{2-16}
     & \multirow{2}{*}{InfoVQA} & Avg. Acc. $\uparrow$ & 96.49 & 83.28 & 53.68 & 58.36 & 61.87 & 81.10 & 80.60 & 51.17 & 58.03 & \underline{81.77} & \textbf{81.94} & 59.20 & 52.34 \\
     &  & Avg. Cost $\downarrow$ & 0.37 & 2.19 & 0.14 & 0.24 & 0.29 & 1.49 & 2.09 & \textbf{0.14} & 0.23 & 1.34 & 1.47 & 0.26 & \underline{0.14} \\
    \cmidrule{2-16}
     & \multirow{2}{*}{MMBench} & Avg. Acc. $\uparrow$ & 98.16 & 89.86 & 69.59 & 69.59 & 81.11 & 85.25 & 70.97 & 62.67 & 69.59 & \textbf{90.32} & \underline{90.32} & 67.74 & 64.52 \\
     &  & Avg. Cost $\downarrow$ & 0.24 & 2.22 & 0.13 & 0.19 & 0.27 & 2.26 & 1.17 & \underline{0.15} & 0.16 & 5.49 & 5.49 & 0.17 & \textbf{0.14} \\
    \cmidrule{2-16}
     & \multirow{2}{*}{MMMU} & Avg. Acc. $\uparrow$ & 98.46 & 74.36 & 45.64 & 42.05 & 53.33 & 60.00 & 65.64 & 37.95 & 39.49 & \underline{76.92} & \textbf{77.44} & 41.03 & 37.44 \\
     &  & Avg. Cost $\downarrow$ & 0.64 & 3.24 & 0.17 & 0.41 & 0.43 & 6.95 & 10.59 & \underline{0.19} & 0.28 & 17.30 & 17.60 & 0.24 & \textbf{0.17} \\
    \cmidrule{2-16}
     & \multirow{2}{*}{MMStar} & Avg. Acc. $\uparrow$ & 96.68 & 72.76 & 49.17 & 54.48 & 59.80 & 68.44 & 66.11 & 46.51 & 55.15 & \underline{73.42} & \textbf{74.09} & 51.16 & 46.84 \\
     &  & Avg. Cost $\downarrow$ & 0.36 & 2.29 & 0.14 & 0.25 & 0.29 & 2.85 & 1.68 & 0.20 & 0.20 & 8.08 & 7.93 & \underline{0.17} & \textbf{0.15} \\
    \cmidrule{2-16}
     & \multirow{2}{*}{RealWorldQA} & Avg. Acc. $\uparrow$ & 99.32 & 77.40 & 67.12 & 55.48 & 63.70 & 67.81 & 60.27 & 57.53 & 60.27 & \textbf{78.08} & \underline{75.34} & 59.59 & 62.33 \\
     &  & Avg. Cost $\downarrow$ & 0.22 & 2.20 & 0.13 & 0.14 & 0.18 & 1.46 & 1.17 & \textbf{0.13} & 0.14 & 3.42 & 3.53 & 0.14 & \underline{0.13} \\
     \midrule
    \multirow{8}{*}{\rotatebox{90}{STEM}} & \multirow{2}{*}{AI2D} & Avg. Acc. $\uparrow$ & 99.69 & 88.21 & 72.33 & 67.45 & 73.58 & 80.66 & 76.42 & 66.20 & 67.30 & \textbf{87.11} & \underline{86.95} & 68.08 & 67.45 \\
     &  & Avg. Cost $\downarrow$ & 0.24 & 2.19 & 0.13 & 0.16 & 0.19 & 2.05 & 2.56 & \textbf{0.13} & 0.14 & 7.02 & 7.09 & 0.14 & \underline{0.13} \\
    \cmidrule{2-16}
     & \multirow{2}{*}{MathVerse} & Avg. Acc. $\uparrow$ & 90.36 & 49.74 & 23.05 & 59.77 & \underline{67.97} & \textbf{68.75} & 66.93 & 46.22 & 56.25 & 65.62 & 64.19 & 43.49 & 25.13 \\
     &  & Avg. Cost $\downarrow$ & 1.06 & 4.67 & 0.18 & 1.64 & 1.70 & 4.01 & 3.19 & 1.03 & 1.35 & 7.56 & 6.75 & \underline{0.96} & \textbf{0.28} \\
    \cmidrule{2-16}
     & \multirow{2}{*}{MathVision} & Avg. Acc. $\uparrow$ & 89.55 & 38.81 & 13.43 & 28.36 & 23.88 & 35.82 & 46.27 & 31.34 & 29.85 & \textbf{61.19} & \underline{61.19} & 22.39 & 22.39 \\
     &  & Avg. Cost $\downarrow$ & 1.77 & 8.16 & 0.17 & 1.23 & \underline{0.72} & 13.10 & 5.64 & 1.00 & 1.04 & 26.73 & 26.73 & 0.84 & \textbf{0.54} \\
    \cmidrule{2-16}
     & \multirow{2}{*}{MathVista} & Avg. Acc. $\uparrow$ & 95.85 & 75.13 & 51.30 & 53.89 & 70.98 & \textbf{77.20} & 73.58 & 57.51 & 55.44 & \underline{77.20} & 77.20 & 49.22 & 50.78 \\
     &  & Avg. Cost $\downarrow$ & 0.36 & 5.00 & 0.16 & 0.50 & 0.54 & 3.46 & 2.81 & 0.67 & 0.42 & 9.63 & 9.63 & \underline{0.30} & \textbf{0.19} \\
     \midrule
    \multirow{8}{*}{\rotatebox{90}{Chart \& Doc}} & \multirow{2}{*}{ChartQA} & Avg. Acc. $\uparrow$ & 97.80 & 86.20 & 81.00 & \textbf{81.40} & 80.80 & \underline{81.00} & 40.60 & 80.00 & 80.00 & 79.20 & 80.60 & 80.40 & 80.60 \\
     &  & Avg. Cost $\downarrow$ & 0.21 & 2.19 & 0.14 & \textbf{0.14} & 0.17 & 1.09 & 0.69 & 0.16 & 0.16 & 2.08 & 2.04 & 0.16 & \underline{0.14} \\
    \cmidrule{2-16}
     & \multirow{2}{*}{DocVQA} & Avg. Acc. $\uparrow$ & 98.56 & 91.48 & 78.95 & 78.66 & 80.48 & 85.55 & 82.30 & 75.50 & 78.47 & \underline{88.23} & \textbf{89.76} & 78.76 & 77.61 \\
     &  & Avg. Cost $\downarrow$ & 0.18 & 2.18 & 0.14 & 0.15 & 0.18 & 1.22 & 1.25 & \textbf{0.14} & 0.15 & 1.12 & 1.29 & 0.16 & \underline{0.14} \\
    \cmidrule{2-16}
     & \multirow{2}{*}{OCRBench} & Avg. Acc. $\uparrow$ & 99.06 & 89.20 & 80.28 & 77.00 & 81.22 & 85.92 & 79.34 & 75.59 & 75.12 & \textbf{92.49} & \underline{92.49} & 77.00 & 76.53 \\
     &  & Avg. Cost $\downarrow$ & 0.23 & 2.41 & 0.14 & 0.16 & 0.19 & 1.44 & 0.96 & \underline{0.15} & 0.17 & 1.58 & 1.68 & 0.15 & \textbf{0.14} \\
    \cmidrule{2-16}
     & \multirow{2}{*}{TextVQA} & Avg. Acc. $\uparrow$ & 89.46 & 73.85 & 72.13 & 70.98 & \textbf{72.13} & 67.05 & 66.09 & 70.69 & \underline{71.17} & 68.58 & 68.68 & 69.83 & 69.83 \\
     &  & Avg. Cost $\downarrow$ & 0.18 & 2.15 & 0.13 & 0.14 & 0.16 & 1.12 & 1.81 & \textbf{0.13} & \underline{0.13} & 1.66 & 1.81 & 0.14 & 0.13 \\
    \midrule
    \multirow{3}{*}{\rotatebox{90}{Average}} & \multirow{3}{*}{All Datasets} & Avg. Acc. $\uparrow$ & 95.60 & \underline{78.01} & 62.43 & 66.26 & 70.68 & 75.49 & 70.01 & 62.92 & 65.93 & 77.49 & \textbf{78.23} & 63.87 & 60.43 \\
    &  & Avg. Cost $\downarrow$ & 0.37 & 2.72 & 0.14 & 0.38 & 0.41 & 2.18 & 2.21 & 0.28 & 0.33 & 3.49 & 3.46 & \underline{0.28} & \textbf{0.16} \\
    &  & Rank Score $\uparrow$ & 93.68 & 68.88 & 64.63 & 67.13 & \textbf{71.09} & \underline{68.92} & 64.62 & 64.41 & 67.04 & 65.61 & 66.56 & 65.33 & 62.58 \\
    \bottomrule
    \end{tabular}
    }
\end{table*}

\end{document}